\documentclass[11pt]{article}

\usepackage[final]{acl}

\usepackage{times}
\usepackage{latexsym}

\usepackage{xcolor}
\usepackage{colortbl}
\newcommand{\best}[1]{\cellcolor{blue!20}{#1}}

\usepackage[T1]{fontenc}

\usepackage[utf8]{inputenc}

\usepackage{microtype}

\usepackage{inconsolata}

\usepackage{graphicx}
\graphicspath{{figures/}}

\usepackage{booktabs}
\usepackage{multirow}

\usepackage{amsmath}
\usepackage{amssymb}

\usepackage{enumitem}

\usepackage{float}

\usepackage{xurl}

\usepackage{listings}
\newcommand{\ours}{\textbf{SPIA}}

\title{Subject-level Inference for Realistic Text Anonymization Evaluation}

\author{
  Myeong Seok Oh$^{1,2}$ \quad Dong-Yun Kim$^{3}$ \quad Hanseok Oh$^{4}$ \quad Chaean Kang$^{3}$ \\
  \textbf{Joeun Kang$^{3}$ \quad Xiaonan Wang$^{3}$ \quad Hyunjung Park$^{1}$ \quad Young Cheol Jung$^{1}$} \\
  \textbf{Hansaem Kim$^{3}$\thanks{Corresponding author.}} \\[0.5em]
  $^{1}$Tscientific, South Korea \quad $^{2}$Soongsil University, South Korea \\
  $^{3}$Yonsei University, South Korea \quad $^{4}$Mila, Canada
}

\begin{document}
\maketitle

\begin{abstract}
Current text anonymization evaluation relies on span-based metrics that fail to capture what an adversary could actually infer, and assumes a single data subject, ignoring multi-subject scenarios. To address these limitations, we present \textbf{SPIA} (Subject-level PII Inference Assessment), the first benchmark that shifts the unit of evaluation from text spans to individuals, comprising 675 documents across legal and online domains with novel subject-level protection metrics. Extensive experiments show that even when over 90\% of PII spans are masked, subject-level inference protection drops as low as 33\%, leaving the majority of personal information recoverable through contextual inference. Furthermore, target-subject-focused anonymization leaves non-target subjects substantially more exposed than the target subject. We show that subject-level inference-based evaluation is essential for ensuring safe text anonymization in real-world settings.\footnote{Code and dataset are available at \url{https://github.com/maisonOP/spia.git}}
\end{abstract}

\section{Introduction}
Text anonymization protects individual privacy by modifying textual data to prevent identification \citep{larbi2022anonymization}. The EU General Data Protection Regulation (GDPR) defines personal data as ``any information relating to an identified or identifiable natural person'' \citep{eu2016gdpr}, requiring protection for all individuals whose information appears in a document. As large language models (LLMs) are increasingly trained on massive text corpora \citep{wang2025generalizationvsmemorizationtracing}, robust anonymization techniques have become essential for protecting privacy and enabling safe data sharing \citep{deusser2025survey,monteiro2024patterns}. However, LLMs simultaneously introduce new privacy risks throughout their lifecycle \citep{wang2025llmprivacy}: they can memorize training data \citep{carlini2021extracting,shokri2017membership,lukas2023analyzing} and infer personal attributes from context without prior exposure to specific individuals \citep{staab2024beyond}. These capabilities enable adversaries to extract sensitive information even from seemingly anonymized texts, fundamentally challenging traditional protection approaches.

Current evaluation methods for text anonymization fail to address these emerging threats in two critical ways. First, span-based metrics measure only whether explicit mentions are masked \citep{pilan2022tab, shen2025piibench, beltrame2024redactbuster}, failing to capture inference risks. \citet{staab2025anonymizers} shows 66.3\% of personal attributes remain inferable even after NER-based anonymization---demonstrating that masking alone cannot prevent inference attacks. Second, existing approaches assume a single target subject \citep{pilan2022tab, manzanares2024evaluating, staab2024beyond}, while real-world texts, such as legal judgments, medical records, and online posts, often mention multiple individuals \citep{shen2025piibench}. Current anonymization techniques largely focus on protecting one primary subject, leaving other mentioned individuals inadequately addressed. These limitations fail to capture whether \emph{all} individuals in a document are actually protected.

\begin{figure*}[t]
\centering
\includegraphics[width=0.95\textwidth]{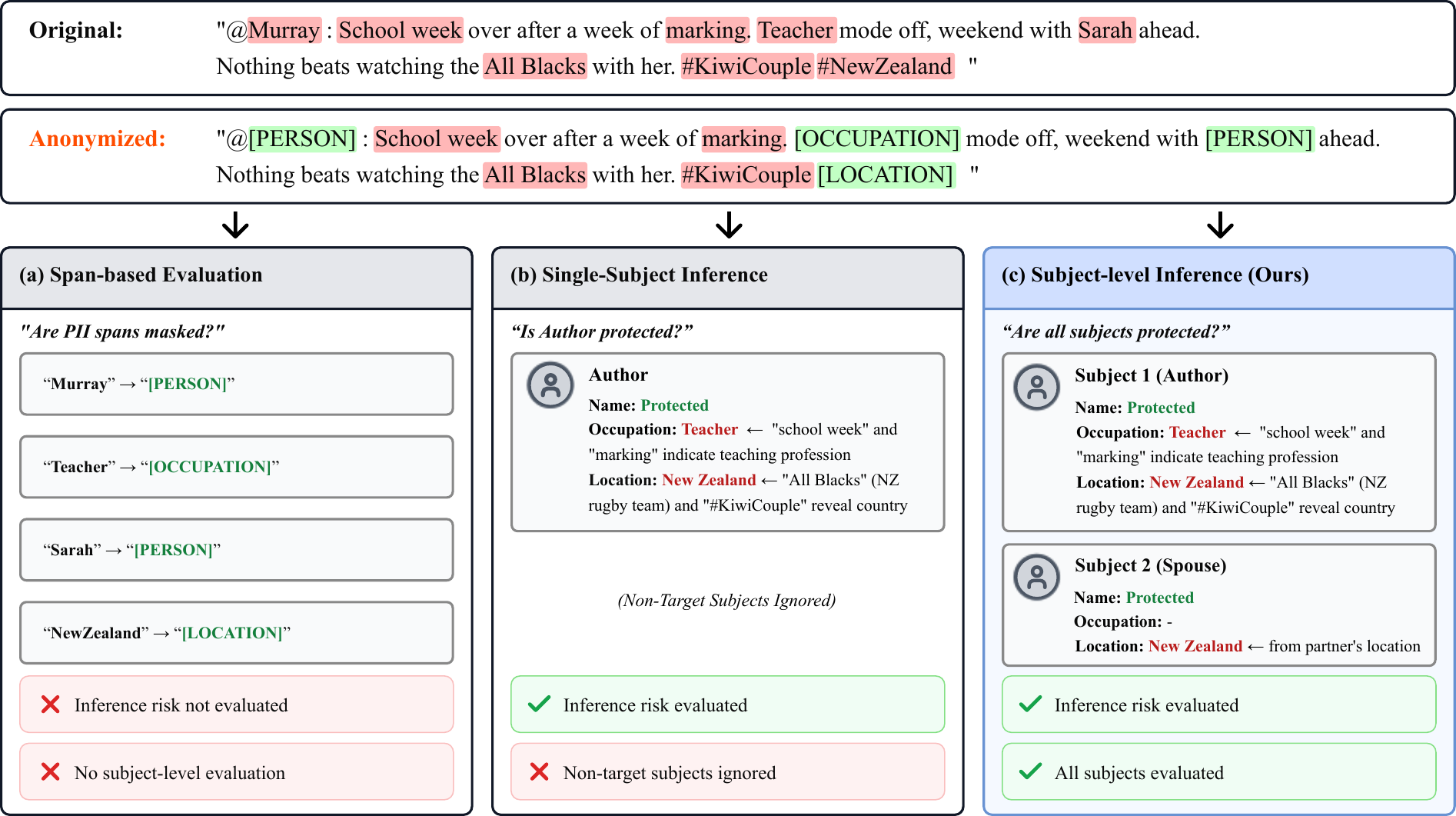}
\caption{Comparison of three evaluation approaches for text anonymization. Span-based evaluation (a) achieves 100\% masking recall but fails to assess whether PII remains inferable from context. Inference-based single-subject evaluation (b) detects inference risk for the target subject, but ignores non-target subjects. Our inference-based subject-level approach (c) evaluates all individuals identifiable from the text, better reflecting real-world scenarios where multiple subjects appear.}
\label{fig:comparison}
\end{figure*}

To address these limitations, we propose \ours~(Subject-level PII Inference Assessment), a benchmark and evaluation framework that shifts the unit of evaluation from text spans to individuals. We define a \emph{subject} as any person identifiable within a document, and treat each subject as the unit of contextual inference, while the ultimate objective is to protect every individual whose information appears in the text. \ours~thus assesses how effectively each individual is actually protected after anonymization. As illustrated in Figure~\ref{fig:comparison}, our subject-level approach evaluates all individuals identifiable from a document, unlike existing span-based or single-subject methods. \ours~comprises 675 documents with 15 PII categories across legal and online domains, and employs a two-stage methodology that identifies all data subjects within a document and infers PII for each subject separately. We introduce novel metrics for per-subject and collective protection assessment. Through extensive experiments across 4 anonymization methods and 6 LLM backbones, we reveal three key findings: first, span-based metrics significantly overestimate protection---despite over 90\% masking rates, inference-based protection remains as low as 33\%; second, anonymization focusing on a target subject can leave non-target subjects less protected; third, these patterns vary substantially across document types, requiring domain-aware approaches.

Our contributions can be summarized as follows:

\begin{itemize}[leftmargin=*]
\item \textbf{\ours~Benchmark}: The first multi-subject, multi-domain benchmark for subject-level privacy assessment in text anonymization.

\item \textbf{Subject-wise Evaluation Framework}: A two-stage methodology with novel metrics for subject-level protection assessment.

\item \textbf{Empirical Findings}: Evidence that span-based metrics overestimate protection, that single-subject-focused anonymization can fail to adequately protect non-target subjects, and that effectiveness varies across document types.
\end{itemize}

% Placeholder for remaining sections
% Sections 2-6 will be added incrementally

\section{Related Work}

\subsection{Personal Data and PII}

Major privacy regulations define personal data scope: GDPR covers information relating to identifiable persons \citep{eu2016gdpr}, while CCPA extends to information reasonably linkable to consumers \citep{ccpa2018}. Prior research classifies PII into direct identifiers (e.g., names) that enable immediate identification, and quasi-identifiers (e.g., date of birth) that enable re-identification when combined \citep{elliot2016anonymisation,domingo2016database}. Notably, gender, birth date, and zip code alone can identify 63--87\% of the U.S. population \citep{sweeney2000simple,golle2006revisiting}, with such risks persisting even in incomplete datasets \citep{rocher2019estimating}. Both thus require protection under privacy regulations \citep{pilan2022tab}.

However, this direct/quasi-identifier distinction is context-dependent, since date of birth is typically a quasi-identifier but can serve as a direct identifier within small groups \citep{pilan2022tab}. We therefore adopt a classification based on structural characteristics: CODE types have fixed formats (e.g., phone numbers, emails), while NON-CODE types are free-text (e.g., names, age). 
\subsection{Text Anonymization}

Text anonymization modifies textual data to protect individual privacy through techniques such as suppression, perturbation, and substitution \citep{larbi2022anonymization}, providing stronger protection than de-identification \citep{kanwal2023balancing}.
A common approach identifies PII spans and masks them, using either NER models \citep{lison2021anonymisation,pilan2022tab} or LLMs instructed to detect and redact sensitive information \citep{liu2023deidgpt}.
Differential privacy has also been applied to text anonymization, framing anonymization as a randomized transformation that limits the distinguishability of texts across individuals \citep{dwork2006calibrating,utpala2023locally}.
Beyond span-level masking, adversarial approaches defend against inference attacks by misleading adversaries or iteratively removing revealing cues \citep{frikha2024incognitext,staab2025anonymizers}.
Recent work has also emphasized evaluating privacy-utility tradeoffs, as overly aggressive anonymization can render text unusable \citep{chen2024mass,yang2025rupta}.
We evaluate anonymization methods from these approaches in Section~\ref{sec:experiments}.

\subsection{Anonymization Evaluation Benchmarks}

Various benchmarks have been proposed for text anonymization. i2b2/UTHealth \citep{stubbs2015annotating} focuses on medical records, WikiPII \citep{hathurusinghe2021privacy} on Wikipedia biographies, and the Text Anonymization Benchmark (TAB) \citep{pilan2022tab} on legal documents with comprehensive PII coverage. PersonalReddit \citep{staab2024beyond} introduced inference-based evaluation for author profiling, while PANORAMA \citep{selvam2025panorama} provides large-scale synthetic data across multiple locales. PII-Bench \citep{shen2025piibench} addresses multi-subject scenarios with query-aware privacy protection evaluation.

We identify four desirable properties for text anonymization benchmarks: (1) \textbf{Coverage}---comprehensive PII types from direct to quasi-identifiers; (2) \textbf{Inference}---addressing context-inferable information beyond explicit mentions; (3) \textbf{Multi-domain}---diverse text domains; (4) \textbf{Subject-aware}---per-subject evaluation in multi-subject scenarios.

\begin{table}[t]
\centering
\small
\begin{tabular}{lccccc}
\toprule
\textbf{Dataset} & \textbf{Scale} & \textbf{Cov.} & \textbf{Inf.} & \textbf{M-D} & \textbf{S-A} \\
\midrule
i2b2/UTHealth & 1,304 & $\triangle$ & \texttimes & \texttimes & \texttimes \\
WikiPII & 23,090 & $\triangle$ & \texttimes & \texttimes & \texttimes \\
TAB & 1,268 & \checkmark & \texttimes & \texttimes & $\triangle$ \\
PersonalReddit & 520 & \texttimes & \checkmark & \texttimes & $\triangle$ \\
PANORAMA & 384K & \checkmark & \texttimes & $\triangle$ & $\triangle$ \\
PII-Bench & 2,842 & \checkmark & \texttimes & \texttimes & \checkmark \\
\ours~(Ours) & 675 & \checkmark & \checkmark & \checkmark & \checkmark \\
\bottomrule
\end{tabular}
\caption{Comparison of existing text anonymization benchmark datasets. Cov.=Coverage, Inf.=Inference, M-D=Multi-domain, S-A=Subject-aware. \checkmark=supported, $\triangle$=limited support, \texttimes=not supported.}
\label{tab:benchmark-comparison}
\end{table}

As shown in Table~\ref{tab:benchmark-comparison}, existing datasets satisfy only some of these properties. TAB achieves broad coverage but lacks inference-based evaluation. PersonalReddit supports inference but targets only the single author. PII-Bench distinguishes subjects but remains at span-based evaluation. \ours~ is the first benchmark satisfying all four properties.

\subsection{Anonymization Evaluation Metrics}
\label{sec:eval-metrics}

\textbf{Token Recall} measures the ratio of masked tokens among all PII tokens \citep{lison2021anonymisation}. \textbf{Entity Recall} measures the ratio of fully protected entities, where an entity is considered protected only when all its occurrences are masked~\citep{pilan2022tab}. It is reported separately for direct identifiers ($ER_{di}$), which enable re-identification individually, and quasi-identifiers ($ER_{qi}$), which enable re-identification only through combination. These span-based metrics evaluate only explicit mentions and cannot capture what an adversary could infer through context. \citet{staab2024beyond}'s \textbf{Adversarial Accuracy (AAC)} measures inference-based privacy risks using LLM adversaries, but assumes a single subject as the protection target. To address these limitations, we propose \textbf{Individual Protection Rate (IPR)} and \textbf{Collective Protection Rate (CPR)}, detailed in Section~\ref{sec:eval-framework}.

% Placeholder for Section 3
\section{SPIA Benchmark Construction}
\label{sec:benchmark}

\begin{figure*}[t]
\centering
\includegraphics[width=0.95\textwidth]{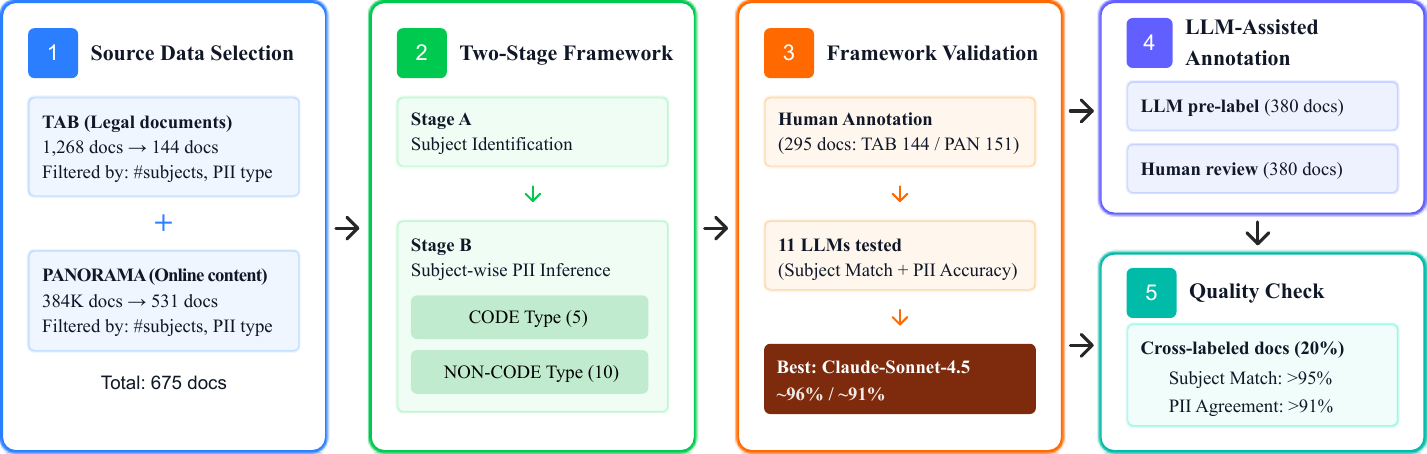}
\caption{\ours~benchmark construction pipeline. Documents are filtered by subject count distribution and PII density to ensure diverse evaluation scenarios. The two-stage framework identifies all subjects (Stage A), then infers CODE (5 types) and NON-CODE (10 types) PIIs per subject (Stage B). After validating 11 LLMs on human-annotated test set, best-performing model pre-labels remaining documents for human review.}
\label{fig:pipeline}
\end{figure*}

\ours~is built on two English text datasets, legal documents (TAB) and online content (PANORAMA), comprising 675 documents with 1,712 subjects and 7,040 PIIs annotated across 15 categories. Figure~\ref{fig:pipeline} illustrates our five-stage construction pipeline.

\subsection{Source Data}

\noindent\textbf{TAB} comprises 144 ECHR legal judgments filtered from 1,268 documents \citep{pilan2022tab} to ensure: (1) varied subject counts from 2 to 5 or more, (2) rich demographic PIIs from TAB's annotations, and (3) test set scale aligned with prior text anonymization studies \citep{papadopoulou2023neural, pilan2025truthful}. These legal records feature applicants, defendants, witnesses, and judges, maintaining consistent facts about each individual throughout---enabling reliable inference-based evaluation.

\noindent\textbf{PANORAMA} includes 531 synthetic online texts from 384,789 documents \citep{selvam2025panorama}. We prioritize documents with diverse subject counts from 1 to 5 or more that contain CODE PIIs. The original dataset has been constructed with enforced attribute consistency (e.g., realistic age gaps in family relationships, coherent education-occupation combinations), making it suitable for inference-based evaluation. We sample 151 documents (~30\%) as the test set, comparable in scale to TAB.

\subsection{Annotation Schema}
\label{sec:annotation-schema}

\noindent\textbf{Subject Identification.} A subject is any individual person whose PII can potentially be inferred from the text. Two key rules are applied for checking potential identification: (1) the same person mentioned multiple times counts as one subject, and (2) collective references (e.g., ``citizens of LA'') are excluded unless a specific count is given (e.g., ``2 citizens''). This approach enables quantifiable privacy risk assessment by restricting our scope to enumerable individuals.

\noindent\textbf{PII Taxonomy.} We define 15 PII categories classified by structural characteristics into CODE and NON-CODE types. CODE types are identifiers with fixed structural patterns: ID Number, Driver License, Phone, Passport, and Email, selected from commonly addressed PII in prior research \citep{fei2024kdpii,selvam2025panorama}. NON-CODE types are free-text or categorical values: Name, Sex, Age, Location, Nationality, Education, Relationship, Occupation, Affiliation, and Position, based on \citet{staab2024beyond}'s classification with additions from existing benchmarks \citep{pilan2022tab,fei2024kdpii}. We include CODE types in inference-based evaluation because pattern-based NER detectors may miss formats unseen during training (e.g., ``(555) 123-4567'' vs.\ ``5551234567'' for a phone number), whereas inference-based approaches can recognize the underlying information regardless of surface form.

\noindent\textbf{Hardness \& Certainty.} Each PII is assessed on inference difficulty (Hardness, 1--5) and confidence (Certainty, 1--5), adopting the schema established in prior inference-based benchmarks~\citep{staab2024beyond,yukhymenko_synthetic_2024}. Hardness represents cognitive effort required for inference, while Certainty indicates confidence based on textual evidence. Detailed annotation guidelines are provided in Appendix~\ref{sec:appendix-g}, with scale examples in Appendix~\ref{sec:appendix-a}.

\subsection{Annotation Process}
\label{sec:annotation-process}

\noindent\textbf{Two-Stage Framework.} \ours~extends \citet{staab2024beyond}'s author profiling approach to subject-level inference. Stage A identifies all subjects in the text with distinguishing descriptions (names, roles, etc.). Stage B infers PIIs for each identified subject, split into separate CODE and NON-CODE calls. This separation (1) avoids requiring the model to identify all 15 categories simultaneously, (2) reduces prompt length, and (3) enables type-appropriate handling---the two types can differ in output format, validation criteria, and annotation requirements, making them better suited to dedicated prompts. The prompts used for each stage are provided in Appendix~\ref{sec:appendix-g2}.

\noindent\textbf{Framework Validation.} To validate this framework, we evaluate 11 LLMs on manually constructed test sets from selected source data. Extending \citet{staab2024beyond}'s methodology, we evaluate both subject matching and PII inference (see Appendix~\ref{sec:appendix-c3} for details). We measure two metrics: Subject Match Ratio for subject identification and Inference Accuracy for PII inference. Claude-Sonnet-4.5 achieves the best performance: Subject Match ~96\% and Inference Accuracy ~91\% on both datasets, and is selected for pre-labeling. Detailed results are in Appendix~\ref{sec:appendix-b}.

\noindent\textbf{Annotation Procedure.} Annotation proceeds in three stages: (1) human annotation for test set (295 docs), (2) LLM pre-labeling using Claude-Sonnet-4.5 for remaining PANORAMA (380 docs), and (3) human review and correction to complete all 675 documents.

\subsection{Quality Control and Dataset Statistics}
\label{sec:statistics}

Inter-annotator agreement (IAA) is measured to verify annotation reliability. Five annotators received overlapping assignments of ~20\% documents from each dataset. Agreement is evaluated in two stages---subject matching between annotator pairs and PII value comparison---using the same scoring scheme as Section~\ref{sec:eval-protocol} with additional human verification. As shown in Table~\ref{tab:iaa}, labelers identify the same subjects in >94\% of cases (TAB: 96.8\%, PANORAMA: 94.7\%) and assign matching PII values in >90\% of comparisons (TAB: 91.3\%, PANORAMA: 94.3\%), with complete disagreement below 4\%. Disagreements are adjudicated through consensus.

\begin{table}[t]
\centering
\small
\begin{tabular}{lcc}
\toprule
\textbf{Metric} & \textbf{TAB} & \textbf{PANORAMA} \\
\midrule
Cross-labeled Docs & 28 & 113 \\
Annotated Subjects & 95 & 222 \\
Subject Match Rate & 96.8\% & 94.7\% \\
Total PII Comparisons & 516 & 634 \\
Match & 91.3\% & 94.3\% \\
Less Precise & 4.8\% & 1.7\% \\
Mismatch & 3.9\% & 3.9\% \\
\textbf{Mean Score} & \textbf{93.7\%} & \textbf{95.2\%} \\
\bottomrule
\end{tabular}
\caption{Subject-wise Inference Inter-Annotator Agreement (IAA) Results. Match/Less Precise/Mismatch indicate PII value agreement levels.}
\label{tab:iaa}
\end{table}

The final \ours~benchmark comprises 675 documents, 1,712 subjects, and 7,040 PIIs across two complementary datasets, as shown in Table~\ref{tab:basic-stats}. 85.7\% (6,033) of all PIIs have Certainty $\geq$ 3; additional dataset details are in Appendix~\ref{sec:appendix-a}.

\begin{table}[t]
\centering
\small
\resizebox{\columnwidth}{!}{%
\begin{tabular}{lccc}
\toprule
\textbf{Metric} & \textbf{TAB} & \textbf{PANORAMA} & \textbf{Total} \\
\midrule
Documents & 144 & 531 & 675 \\
Num. of Subjects & 586 & 1,126 & 1,712 \\
Avg Subjects/Doc & 4.07 & 2.12 & 2.54 \\
\shortstack[l]{Num. of PIIs\\[1.65ex]~} & \shortstack[c]{3,350\\(3,064)} & \shortstack[c]{3,690\\(2,969)} & \shortstack[c]{7,040\\(6,033)} \\
Avg PIIs/Subject & 5.72 & 3.28 & 4.11 \\
Avg Doc Length (chars) & 3,918 & 260 & - \\
\bottomrule
\end{tabular}%
}
\caption{\ours~Dataset Basic Statistics. Numbers in parentheses indicate PIIs with Certainty $\geq$ 3. TAB contains longer legal documents with more subjects per document, while PANORAMA offers shorter online texts with diverse PII types including CODE PIIs.}
\label{tab:basic-stats}
\end{table}

\section{Evaluation Framework}
\label{sec:eval-framework}

We propose a subject-level evaluation framework with novel metrics that capture inferable privacy risks across all subjects.

\subsection{Subject-level Privacy Metrics}

Let $N$ be the total number of subjects in a document, $O_i$ be the number of Ground Truth PIIs for subject $i$ in the original text, and $A_i$ be the number of PIIs an adversary can still infer from the anonymized text for subject $i$.

\noindent\textbf{Collective Protection Rate (CPR)} measures the proportion of protected PIIs across all subjects, where subjects with more PIIs naturally contribute more to the overall score:
\begin{equation}
\text{CPR} = 1 - \frac{\sum_{i=1}^{N} A_i}{\sum_{i=1}^{N} O_i}
\end{equation}

\noindent\textbf{Individual Protection Rate (IPR)} is the average of per-subject protection rates, assigning equal weight to all subjects regardless of their PII count:
\begin{equation}
\text{IPR} = \frac{1}{N} \sum_{i=1}^{N} \left(1 - \frac{A_i}{O_i}\right)
\end{equation}

For both metrics, 1 indicates full protection and 0 indicates full exposure. A concrete calculation example is provided in Appendix~\ref{sec:cpr-ipr-example}.

\subsection{Evaluation Protocol}
\label{sec:eval-protocol}

Computing CPR and IPR requires subject alignment between Ground Truth annotations and adversarial inference results from anonymized text, along with PII-level comparison. We define a 3-step evaluation pipeline.

\noindent\textbf{Step 1: Subject Matching.} For each document, establish one-to-one correspondence between Ground Truth subjects from original text and subjects identified from anonymized text. The correspondence is determined based on subject descriptions and contextual information. Unmatched Ground Truth subjects are assigned 0 points for all PIIs.

\noindent\textbf{Step 2: PII Scoring.} For matched subject pairs, apply \citet{staab2024beyond}'s scoring scheme to compare individual PIIs: 1.0 for exact match, 0.5 for partial match (e.g., inferring ``California'' for Ground Truth ``Los Angeles''), 0.0 for mismatch.

\noindent\textbf{Step 3: Metric Calculation.} Compute CPR and IPR from the aggregated scores. Detailed implementation of each step is described in Appendix~\ref{sec:appendix-c}.

\section{Experiments}
\label{sec:experiments}

\subsection{Experimental Setup}

\textbf{Evaluation Procedure.} The experiments follow a three-phase process: \textbf{(1) Anonymization}---original texts are anonymized using each method and backbone combination, with TAB Longformer using a single model while other methods use 6 backbones (GPT-4.1, GPT-4.1-Mini, Claude-Sonnet-4.5, Claude-Haiku-4.5, Llama-3.1-8B, Gemma-3-27B), generating 19 configurations in total. \textbf{(2) Subject-wise PII Inference}---the adversarial LLM (Claude-Sonnet-4.5) applies the two-stage framework from Section~\ref{sec:annotation-process} to identify subjects and infer 15 PII categories from the anonymized outputs. \textbf{(3) Evaluation}---we assess each method from three perspectives:
\begin{itemize}[leftmargin=*]
\item \textbf{Span-based Evaluation}: Measures whether Ground Truth PII spans are masked in anonymized text. Token Recall ($R_{di+qi}$) evaluates at the individual mention level, while Entity Recall evaluates based on whether all spans of the same entity are masked, separately for direct identifiers ($ER_{di}$) and quasi-identifiers ($ER_{qi}$).
\item \textbf{Inference-based Evaluation}: Measures whether PII can still be inferred from anonymized text. After matching Ground Truth subjects from original text with subjects identified from anonymized text, PII-level scoring is performed to calculate CPR and IPR as defined in Section~\ref{sec:eval-framework}. Additionally, 1-AAC is reported to express AAC as a protection rate, which measures protection for the target subject only (applicant for TAB, author for PANORAMA).
\item \textbf{Utility Evaluation}: We adopt \citet{staab2024beyond}'s methodology, computing Mean Utility as the average of LLM-based Readability, Meaning, and ROUGE-L scores.
\end{itemize}

Details on backbones and the evaluation procedure are described in Appendix~\ref{sec:appendix-c}. To verify that evaluation outcomes are robust to adversary choice, we additionally vary the adversary across GPT-4.1 and Claude-Haiku-4.5, obtaining Spearman $\rho > 0.98$ for both CPR and IPR across all anonymization configurations (see Appendix~\ref{sec:multi-adversary-full}).

\begin{table*}[t]
\centering
\fontsize{7.5}{9}\selectfont
\setlength{\tabcolsep}{3pt}
\renewcommand{\arraystretch}{0.82}
\begin{tabular}{llccccccccccccccc}
\toprule
& & \multicolumn{7}{c}{\textbf{(a) PANORAMA (N=151)}} & & \multicolumn{7}{c}{\textbf{(b) TAB (N=144)}} \\
\cmidrule(lr){3-9} \cmidrule(lr){11-17}
& & \multicolumn{3}{c}{Span-based} & \multicolumn{3}{c}{Inference-based} & \textbf{Util.} & &
\multicolumn{3}{c}{Span-based} & \multicolumn{3}{c}{Inference-based} & \textbf{Util.} \\
\cmidrule(lr){3-5} \cmidrule(lr){6-8} \cmidrule(lr){9-9}
\cmidrule(lr){11-13} \cmidrule(lr){14-16} \cmidrule(lr){17-17}
\textbf{Method} & \textbf{Backbone}
& $R$ & $ER_{di}$ & $ER_{qi}$ & 1-AAC & CPR & IPR & Mean
& & $R$ & $ER_{di}$ & $ER_{qi}$ & 1-AAC & CPR & IPR & Mean \\
\midrule
Longformer & -- 
& .883 & .873 & .716 & .589 & .597 & .585 & .820
& & .940 & .997 & .923 & .384 & .330 & .325 & .874 \\
\midrule
\multirow{6}{*}{DeID-GPT}
& Llama-3.1-8B
& .958 & .997 & .865 & \best{.819} & .840 & .866 & .934
& & .889 & \best{1.00} & .895 & .495 & .396 & .388 & \best{.961} \\
& Gemma-3-27B
& .944 & .991 & .791 & .679 & .684 & .688 & .935
& & .978 & \best{1.00} & .980 & .536 & .519 & .505 & .872 \\
& GPT-4.1-Mini
& .959 & .997 & .828 & .687 & .687 & .689 & .942
& & .947 & \best{1.00} & .956 & .504 & .430 & .418 & .959 \\
& GPT-4.1
& \best{.984} & \best{1.00} & .921 & .775 & .799 & .820 & .817
& & \best{.990} & \best{1.00} & \best{.991} & \best{.638} & .674 & .665 & .754 \\
& Claude-Haiku
& .921 & .994 & .712 & .695 & .694 & .689 & \best{.946}
& & .972 & \best{1.00} & .974 & .623 & .570 & .553 & .947 \\
& Claude-Sonnet
& .969 & \best{1.00} & .865 & .711 & .727 & .735 & .926
& & .988 & .993 & .990 & .628 & .650 & .635 & .770 \\
\midrule
\multirow{6}{*}{DP-Prompt}
& Llama-3.1-8B
& .719 & .659 & .721 & .467 & .480 & .519 & .634
& & .855 & .810 & .859 & .154 & .578 & .579 & .684 \\
& Gemma-3-27B
& .387 & .303 & .447 & .124 & .144 & .182 & .744
& & .899 & .869 & .919 & .212 & \best{.684} & \best{.689} & .540 \\
& GPT-4.1-Mini
& .276 & .146 & .312 & .131 & .137 & .157 & .843
& & .289 & .356 & .216 & .032 & .132 & .147 & .851 \\
& GPT-4.1
& .208 & .115 & .251 & .138 & .130 & .162 & .833
& & .561 & .353 & .512 & .047 & .137 & .149 & .785 \\
& Claude-Haiku
& .360 & .189 & .419 & .148 & .169 & .200 & .757
& & .782 & .363 & .762 & .081 & .346 & .361 & .755 \\
& Claude-Sonnet
& .388 & .204 & .498 & .151 & .194 & .229 & .772
& & .789 & .450 & .770 & .067 & .452 & .446 & .764 \\
\midrule
\multirow{6}{*}{\shortstack[l]{Adversarial\\Anon.}}
& Llama-3.1-8B
& .890 & .923 & .819 & .701 & .759 & .763 & .825
& & .665 & .619 & .640 & .510 & .511 & .510 & .753 \\
& Gemma-3-27B
& .946 & .991 & .842 & .709 & .799 & .835 & .829
& & .815 & .934 & .797 & .396 & .339 & .341 & .809 \\
& GPT-4.1-Mini
& .953 & .997 & .874 & .737 & .831 & .859 & .844
& & .555 & .955 & .494 & .480 & .432 & .421 & .885 \\
& GPT-4.1
& .969 & .994 & .930 & .795 & \best{.870} & \best{.897} & .820
& & .894 & \best{1.00} & .881 & .450 & .359 & .365 & .857 \\
& Claude-Haiku
& .970 & .997 & .907 & .728 & .789 & .825 & .853
& & .554 & .723 & .519 & .342 & .308 & .307 & .894 \\
& Claude-Sonnet
& .979 & \best{1.00} & \best{.944} & .785 & .852 & .875 & .815
& & .727 & .990 & .717 & .472 & .362 & .364 & .867 \\
\bottomrule
\end{tabular}
\caption{Privacy and Utility Evaluation Results across anonymization methods and datasets. Higher values indicate better protection and utility. Span-based metrics include Token Recall ($R$=$R_{di+qi}$) and Entity Recall ($ER_{di}$, $ER_{qi}$). Inference-based metrics measure protection for target subject (1-AAC) or all subjects (CPR, IPR). Mean Utility combines Readability, Meaning, and ROUGE-L. Blue cells indicate the highest value per metric (ties allowed).}
\label{tab:combined-results}
\end{table*}

\begin{figure*}[t]
\centering
\includegraphics[width=0.80\textwidth]{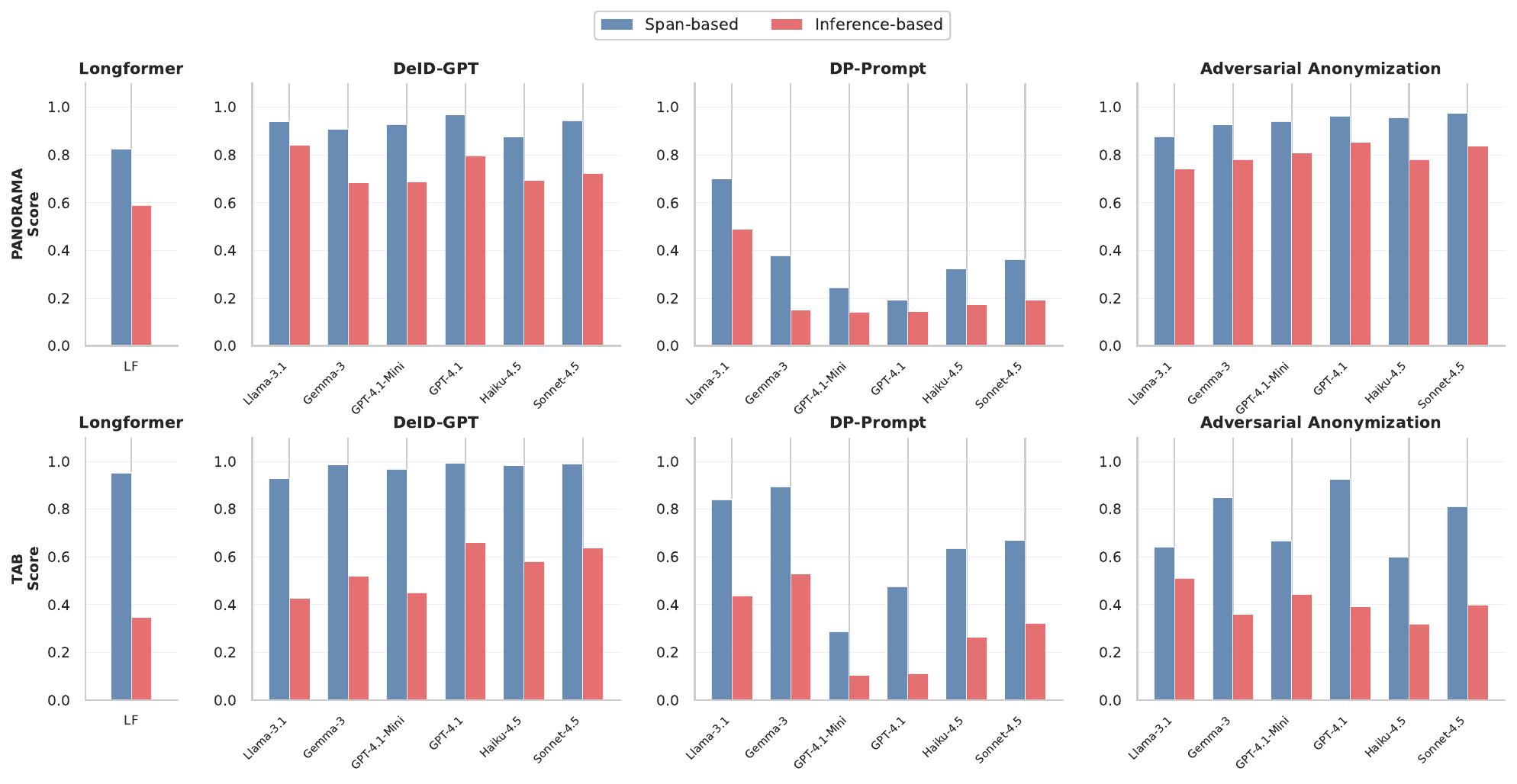}
\caption{Per-backbone comparison of span-based and inference-based metric averages across four anonymization techniques (columns) and two datasets (rows). Span-based metrics consistently exceed inference-based metrics across all configurations, with larger gaps on TAB than PANORAMA. Longformer uses a single model; other methods show results for six LLM backbones.}
\label{fig:span-vs-inference}
\end{figure*}

\begin{figure*}[!t]
\centering
\includegraphics[width=0.80\textwidth]{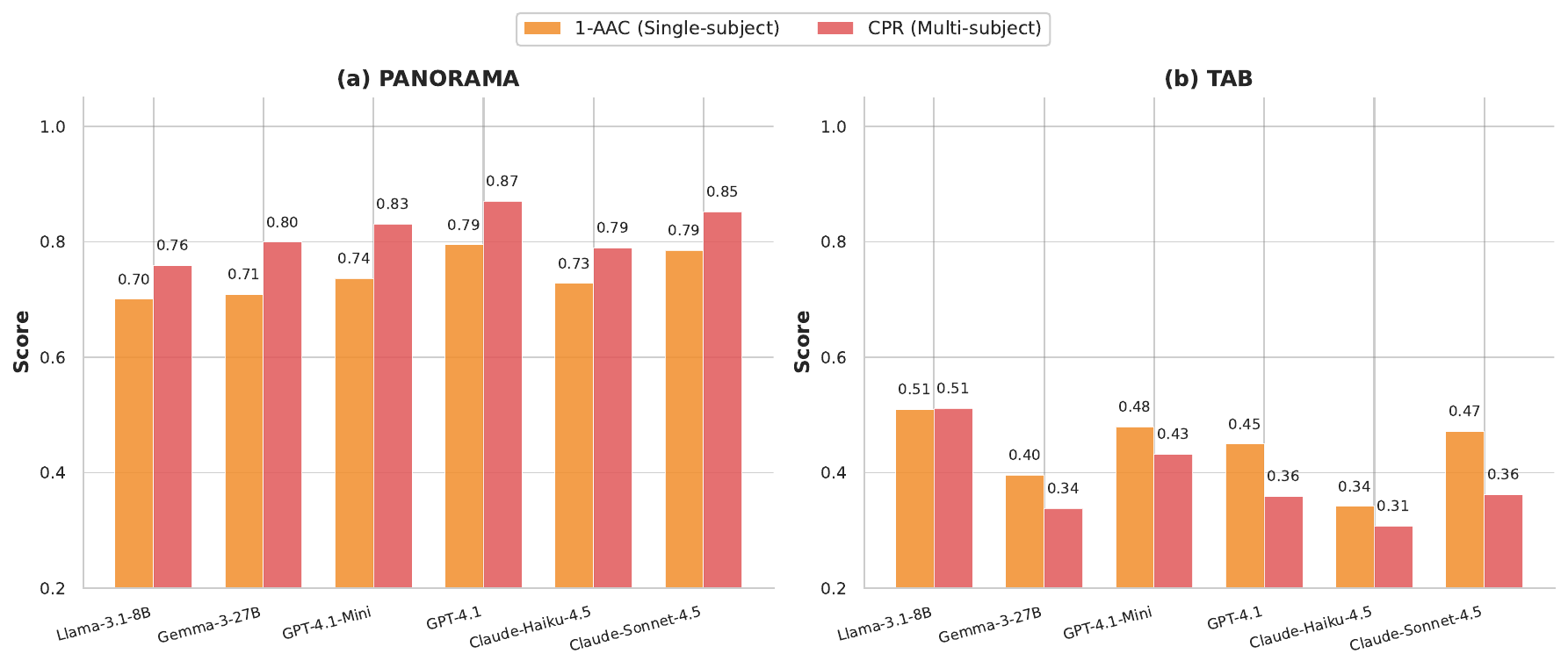}
\caption{Comparison of single-subject (1-AAC) and multi-subject (CPR) metrics for the Adversarial Anonymization technique across 6 LLM backbones. On TAB, 1-AAC exceeds CPR, revealing that non-target subjects are inadequately protected. On PANORAMA, the pattern reverses with CPR exceeding 1-AAC.}
\label{fig:single-vs-multi}
\end{figure*}

\subsection{Anonymization Methods}

Four methods representing major approaches to text anonymization are selected. Detailed method parameters and backbone configurations are described in Appendix~\ref{sec:appendix-c}.

\noindent\textbf{TAB Longformer} \citep{pilan2022tab}: An NER-based token classification approach that identifies tokens corresponding to PII and replaces them with masking tokens (\texttt{[PERSON]}, \texttt{[LOC]}, etc.).

\noindent\textbf{DeID-GPT} \citep{liu2023deidgpt}: A zero-shot prompting-based anonymization technique that focuses on removing explicit PIIs by instructing LLMs to find defined PII categories and replace them with \texttt{[redacted]}.

\noindent\textbf{DP-Prompt} \citep{utpala2023locally}: A method that paraphrases text with high temperature to obfuscate the author's writing style and linguistic patterns.

\noindent\textbf{Adversarial Anonymization (AA)} \citep{staab2024beyond}: An iterative technique that uses an adversarial inference model to identify revealing cues in text, then removes them to prevent personal attribute inference. It is primarily designed to defend against author profiling and focuses on a single subject (in this work, the applicant for TAB and the author for PANORAMA).

\subsection{Results}

Table~\ref{tab:combined-results} presents the privacy and utility evaluation results for 4 anonymization methods and 6 backbones across PANORAMA and TAB datasets.

\noindent\textbf{Method Comparison.}
Longformer achieves near-perfect span masking ($ER_{di}$ 0.997) but the lowest CPR (0.330), demonstrating that entity recognition alone cannot prevent inference attacks.
DeID-GPT, which also targets explicit PII spans but uses LLM-based detection, achieves both high masking ($R_{di+qi}$ up to 0.990) and competitive inference protection (CPR/IPR 0.799/0.820), with the highest utility (up to 0.961), suggesting that LLM-based detection enables more flexible anonymization than pattern-based NER.
DP-Prompt shows the lowest average CPR (0.30), suggesting style-obfuscating differential privacy offers limited inference protection.
AA provides the strongest inference protection (CPR/IPR 0.870/0.897) through iterative adversarial refinement, but with moderate utility trade-off.

\noindent\textbf{Backbone Comparison.}
GPT-4.1 leads in both span-based and inference-based protection, while smaller backbone models show competitive results in specific cases---Llama-3.1-8B achieves the highest 1-AAC (0.819) under DeID-GPT. However, this reflects effective literal category-based masking on short PANORAMA texts rather than superior anonymization capability (see Appendix~\ref{sec:backbone-behavior-analysis}).
For utility, Claude-Haiku achieves strong scores (up to 0.947), while GPT-4.1 shows lower utility despite its strong privacy protection, indicating a privacy-utility trade-off.

\section{Analysis}
\label{sec:analysis}

\subsection{High Span Masking Does Not Guarantee Inference Protection}

As shown in Figure~\ref{fig:span-vs-inference}, span-based metrics are consistently higher than inference-based metrics across all methods and backbones. We identify two key findings: \textbf{(1) The span-inference gap is substantial and universal.} Across all 19 configurations, the gap ranges from 0.10 to 0.61, present even in the best-performing setup (AA with GPT-4.1 on PANORAMA). \textbf{(2) NER-based methods exhibit the largest gap.} Longformer on TAB achieves near-perfect span masking ($ER_{di}$ 0.997) while CPR is only 0.330---two-thirds of PIIs remain inferable despite virtually complete span removal. LLM-based methods generally show smaller gaps on TAB, suggesting instruction-following models better identify contextual cues. These results demonstrate that inference-based metrics are essential for capturing residual privacy risks.

\subsection{Target-Subject-Focused Approaches Underestimate Multi-Subject Privacy Risks}
\label{sec:single-subject-analysis}

On TAB (avg. 4.07 subjects per document), AA shows 1-AAC higher than CPR in most configurations (Figure~\ref{fig:single-vs-multi}). We draw two observations: \textbf{(1) Target-subject-focused anonymization creates protection inequality.} Across 5 of 6 backbones, AA shows 1-AAC exceeding CPR, with up to 11 percentage points gap (Claude-Sonnet: 0.472 vs 0.362), indicating that non-target subjects receive substantially lower protection than the designated target subject. \textbf{(2) Subject-agnostic methods can achieve better collective protection.} DeID-GPT with GPT-4.1 achieves CPR of 0.674, surpassing AA's 0.359 with the same backbone. This counterintuitive result suggests that iterative optimization for the target subject may compromise collective protection by preserving contextual information about non-target subjects. These results highlight the need for anonymization strategies that explicitly protect all subjects present.

\subsection{Anonymization Effectiveness Varies Substantially Across Domains}

Comparing PANORAMA (avg. 260 chars) and TAB (avg. 3,918 chars), we observe domain-dependent patterns: \textbf{(1) The span-inference gap varies by domain characteristics.} The gap is consistently larger on TAB (0.31--0.61) than PANORAMA (0.10--0.29), attributable to TAB containing exclusively NON-CODE PIIs (100\% vs 81\%) where removing inference cues is harder, combined with 15$\times$ longer documents providing more inferential context. Moreover, on TAB, three of four methods fail to protect higher-Hardness PIIs (levels 4--5) as effectively as lower-Hardness ones, while PANORAMA maintains consistent protection across levels. See Appendix~\ref{sec:appendix-e} for further analysis. \textbf{(2) The single-subject vs multi-subject protection pattern reverses across domains.} TAB shows 1-AAC higher than CPR, while PANORAMA shows CPR exceeding 1-AAC by 5.8--9.4 percentage points. This reflects structural differences: TAB's legal documents describe parties independently, while PANORAMA's author-centric content interweaves subjects---anonymizing ``\textit{Married life with Lisa}'' to ``\textit{Life with others}'' inherently protects related individuals. These findings suggest that anonymization effectiveness depends substantially on document characteristics including length, PII distribution, and narrative structure.

\section{Conclusion}

We introduce \ours, the first benchmark for subject-level privacy evaluation in text anonymization, comprising 675 documents with 1,712 subjects across legal and online domains. Through experiments with 4 anonymization methods and 6 LLM backbones, we draw three main findings: (1) High span masking does not guarantee inference protection---even with 99.7\% entity recall, two-thirds of PIIs remain inferable. (2) Single-subject-focused anonymization creates protection inequality, leaving non-target subjects up to 11 percentage points less protected. (3) Anonymization effectiveness varies substantially across domains, requiring domain-aware evaluation approaches.
These findings underscore the necessity of inference-based, multi-subject evaluation frameworks that go beyond span-based metrics. We envision \ours~as a foundation for developing anonymization techniques that ensure equitable privacy protection for all individuals in a document, and release the benchmark and evaluation framework to support continued research in this direction.

\section*{Limitations}

\noindent\textbf{Exclusion of Collective References.} Our annotation focuses on quantifiable individuals, excluding collective references without specified counts (e.g., ``citizens of LA''). Under GDPR \citep{eu2016gdpr}, such group mentions can still enable individual identification when combined with additional context (e.g., in small organizations).

\noindent\textbf{Equal Treatment of PII Risk.} The IPR and CPR metrics proposed in this study treat all PII categories with equal weight. However, actual re-identification risk varies by PII type---for example, a CPR of 0.8 carries substantially different risk if the exposed 20\% consists of direct identifiers (e.g., SSN, passport numbers) versus quasi-identifiers (e.g., gender, age group). Furthermore, combinations of quasi-identifiers (e.g., gender + age + residence) can enable re-identification even when individual PIIs seem benign. Future work could develop metrics that apply risk weights by PII type or model combinatorial re-identification risk from a k-anonymity perspective~\citep{sweeney2002k}. For deployment scenarios, we recommend supplementary analysis stratified by PII sensitivity tiers.

\noindent\textbf{Interdependence of PII Categories and Subject Evaluation.} This study uses 15 PII categories from prior research~\citep{staab2024beyond,pilan2022tab}. However, reducing PII categories could exclude subjects with no remaining inferable PIIs; future work could decouple subject identification from PII-based scoring to address this.

\noindent\textbf{Ambiguity in PII Attribution to Subjects.} Subject-level evaluation requires attributing each PII to a specific subject, which may be ambiguous when context is insufficient. Careful guideline design is needed to ensure consistent attribution across annotators and models.

\noindent\textbf{Scope and Scale.} This study covers two domains, legal documents (TAB) and online content (PANORAMA), across diverse cultural contexts. However, the 675-document scale is constrained by the high cost of subject-level annotation, and only English texts are analyzed. Different languages present unique PII inference pathways: for instance, pro-drop languages such as Korean and Japanese frequently omit subjects, increasing the difficulty of subject boundary detection, while East Asian honorific systems implicitly encode age and social relationships between subjects. Beyond language, extending to other domains would require domain-specific adaptation: clinical notes would require mapping to PHI categories under HIPAA regulations, while audio transcripts would necessitate speaker diarization as a preprocessing step for subject identification.

\section*{Reproducibility}

To ensure the reproducibility of this study, we release the following materials.

\noindent\textbf{Code and Experiment Scripts.} All code, experiment scripts, and configuration files used in this study are publicly available at the following repository: \url{https://github.com/maisonOP/spia.git}. The repository includes (1) implementations of 4 anonymization techniques, (2) subject-level PII inference pipeline, (3) CPR/IPR/1-AAC evaluation scripts, and (4) utility evaluation tools.

\noindent\textbf{Dataset.} We release the \ours~benchmark dataset, comprising TAB (144 documents) and PANORAMA (531 documents), along with per-subject Ground Truth annotations. TAB is derived from \citet{pilan2022tab} (MIT License) and PANORAMA from \citet{selvam2025panorama} (CC BY 4.0 License), with original copyright notices preserved.

\noindent\textbf{Backbone Version Control.} For API-accessed backbones, version control is not fully controllable; we have documented the backbone versions and API call settings at the time of experiments in the Appendix. Open-source backbones (Llama 3.1, Gemma 3, Qwen 3, GPT-OSS) are run locally.

\noindent\textbf{Experimental Settings.} All hyperparameters for anonymization techniques, evaluation settings, and prompt templates are documented in the Appendix. In particular, methods are reproduced as faithfully as possible to the original paper settings, and modifications made in this study (such as applying the TAB 8-category system) are explicitly described.

\section*{Ethical Considerations}

Both datasets used in this study are designed with privacy protection considerations. Annotation was conducted by privacy experts and university researchers from consortium institutions, participating as part of a government-funded research project with compensation covered by the project grant.

\noindent\textbf{PANORAMA} consists entirely of synthetic data, with all profiles and PIIs unrelated to real individuals or actual records. The entire pipeline from profile generation to content generation is composed of model-based generation and constraint-based selection, structurally preventing the possibility of including real personal information \citep{selvam2025panorama}. Therefore, it can be safely used in environments where PII research is needed but use of actual personal information is prohibited.

\noindent\textbf{TAB} is constructed using only judgments for which the European Court of Human Rights (ECHR) legally mandates publication and has received explicit consent from applicants. The ECHR separately de-identifies or excludes from publication sensitive cases or those requiring anonymization before the publication stage, so documents included in TAB have minimized risk of personal information exposure. Additionally, the TAB annotation process provided guidelines to base masking decisions only on publicly inferable information, structurally preventing re-identification possibilities based on non-public information \citep{pilan2022tab}.

\section*{Acknowledgments}

This work was supported in part by the Personal Information Protection Commission and the Korea Internet \& Security Agency (KISA), Republic of Korea, under Project 2780000030; and in part by the Government of the Republic of Korea.

% Bibliography
\bibliography{custom}

\appendix

\section{Dataset Statistics}
\label{sec:appendix-a}

This appendix presents detailed statistics of the \ours~dataset mentioned in Section~\ref{sec:benchmark}.

\subsection{PII Category Distribution}

Table~\ref{tab:pii-category} shows the frequency of each PII category in TAB and PANORAMA. CODE-type PIIs (ID Number, Driver License, Phone, Passport, Email) appear only in PANORAMA, while NON-CODE-type PIIs are distributed across both datasets with different patterns reflecting their domain characteristics.

\begin{table}[t]
\centering
\small
\resizebox{\columnwidth}{!}{%
\begin{tabular}{clccc}
\toprule
\textbf{Type} & \textbf{Category} & \textbf{PANORAMA} & \textbf{TAB} & \textbf{Total} \\
\midrule
\multirow{5}{*}{CODE} & ID Number & 95 & -- & 95 \\
& Driver License & 75 & -- & 75 \\
& Phone & 85 & -- & 85 \\
& Passport & 31 & -- & 31 \\
& Email & 90 & -- & 90 \\
\midrule
\multirow{10}{*}{NON-CODE} & Name & 691 & 366 & 1,057 \\
& Sex & 556 & 392 & 948 \\
& Age & 159 & 153 & 312 \\
& Location & 452 & 510 & 962 \\
& Nationality & 452 & 519 & 971 \\
& Education & 101 & 294 & 395 \\
& Relationship & 254 & 55 & 309 \\
& Occupation & 441 & 453 & 894 \\
& Affiliation & 193 & 292 & 485 \\
& Position & 15 & 316 & 331 \\
\midrule
& \textbf{Total} & \textbf{3,690} & \textbf{3,350} & \textbf{7,040} \\
\bottomrule
\end{tabular}%
}
\caption{PII category frequency distribution across TAB and PANORAMA datasets. TAB is dominated by NON-CODE type PIIs (Name, Location, Occupation), while PANORAMA includes CODE-type PIIs (Email, Phone) that are absent in TAB.}
\label{tab:pii-category}
\end{table}

\subsection{Subject Distribution}

Table~\ref{tab:subject-distribution} shows the distribution of the number of subjects per document. In PANORAMA, documents with 1-2 subjects account for 76.9\%, while in TAB, documents with 5 or more subjects account for 39.6\%, focusing on multi-subject scenarios. The average number of subjects across the entire \ours~dataset is 2.54.

\begin{table}[t]
\centering
\small
\begin{tabular}{cccc}
\toprule
\textbf{\# Subjects} & \textbf{PANORAMA} & \textbf{TAB} & \ours(Total) \\
\midrule
1 & 175 (33.0\%) & - & 175 (25.9\%) \\
2 & 233 (43.9\%) & 22 (15.3\%) & 255 (37.8\%) \\
3 & 58 (10.9\%) & 35 (24.3\%) & 93 (13.8\%) \\
4 & 24 (4.5\%) & 30 (20.8\%) & 54 (8.0\%) \\
5+ & 41 (7.7\%) & 57 (39.6\%) & 98 (14.5\%) \\
\midrule
\textbf{Total} & \textbf{531} & \textbf{144} & \textbf{675} \\
\textbf{Average} & 2.12 & 4.07 & 2.54 \\
\bottomrule
\end{tabular}
\caption{Distribution of Number of Subjects per Document. TAB shows higher subject counts due to multi-party legal proceedings, while PANORAMA covers single to multi-subject online texts.}
\label{tab:subject-distribution}
\end{table}

\subsection{PII per Subject}

Table~\ref{tab:pii-per-subject} shows the distribution of PIIs per subject. TAB contains an average of 5.72 PIIs per subject, providing richer PII information than PANORAMA (3.28). This is because legal documents record detailed personal information of the parties involved.

\begin{table}[t]
\centering
\small
\resizebox{\columnwidth}{!}{%
\begin{tabular}{cccc}
\toprule
\textbf{\# PIIs} & \textbf{PANORAMA} & \textbf{TAB} & \ours(Total) \\
\midrule
1 & 159 (14.1\%) & 10 (1.7\%) & 169 (9.9\%) \\
2 & 229 (20.3\%) & 19 (3.2\%) & 248 (14.5\%) \\
3 & 360 (32.0\%) & 34 (5.8\%) & 394 (23.0\%) \\
4 & 135 (12.0\%) & 46 (7.8\%) & 181 (10.6\%) \\
5 & 93 (8.3\%) & 107 (18.3\%) & 200 (11.7\%) \\
6 & 88 (7.8\%) & 190 (32.4\%) & 278 (16.2\%) \\
7 & 45 (4.0\%) & 107 (18.3\%) & 152 (8.9\%) \\
8+ & 17 (1.5\%) & 73 (12.4\%) & 90 (5.3\%) \\
\midrule
\textbf{Total Subjects} & \textbf{1,126} & \textbf{586} & \textbf{1,712} \\
\textbf{Average} & 3.28 & 5.72 & 4.11 \\
\bottomrule
\end{tabular}%
}
\caption{Distribution of Number of PIIs per Subject. Most subjects have 2--6 inferable PIIs, providing sufficient signal for inference-based evaluation while reflecting realistic privacy exposure scenarios.}
\label{tab:pii-per-subject}
\end{table}

\subsection{Certainty and Hardness Distribution}

\begin{figure*}[t]
\centering
\begin{minipage}{0.47\textwidth}
\centering
\includegraphics[width=\linewidth]{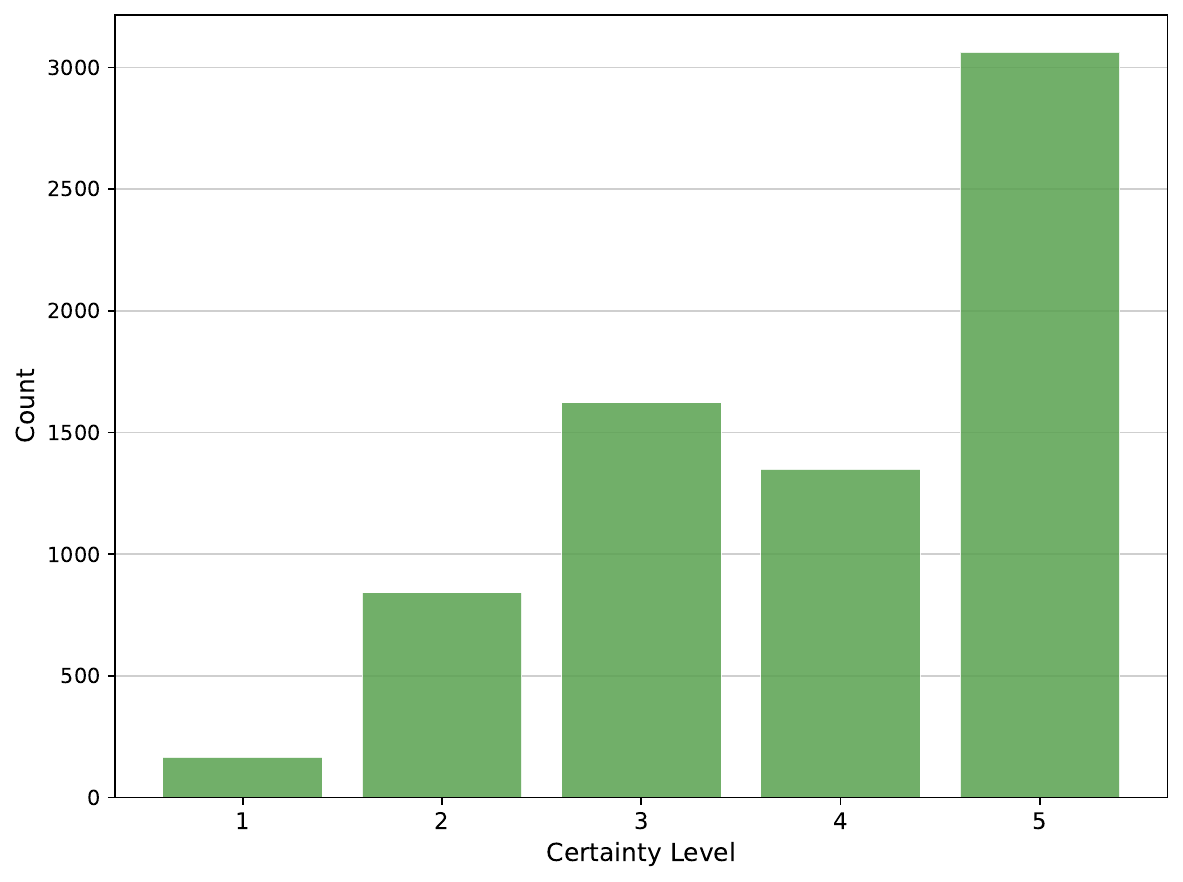}
\caption{PII Certainty distribution. The majority of PIIs (85.7\%) have Certainty $\geq$ 3, meaning they have direct or indirect evidence in the text.}
\label{fig:certainty-distribution}
\end{minipage}
\hfill
\begin{minipage}{0.47\textwidth}
\centering
\includegraphics[width=\linewidth]{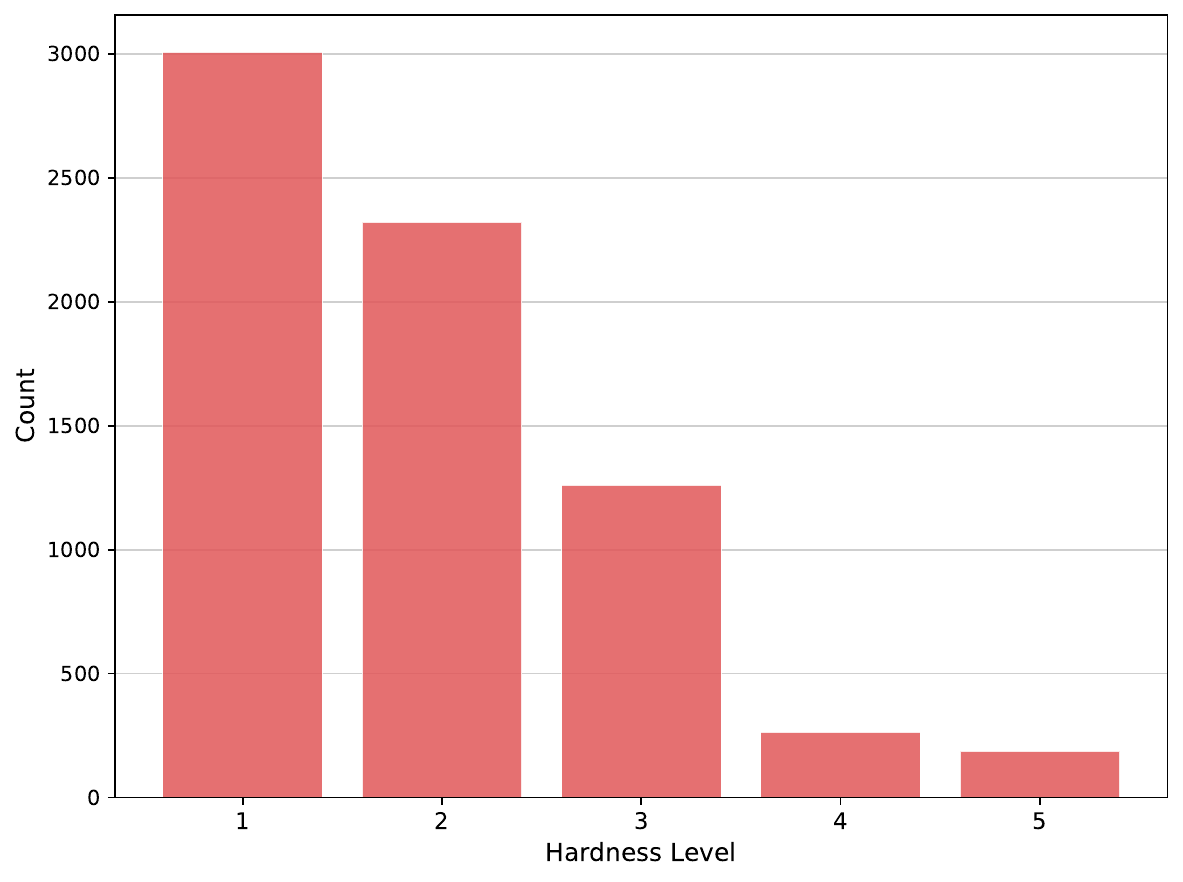}
\caption{PII Hardness distribution. Most PIIs (75.7\%) require low cognitive effort to infer, suggesting that adversaries can easily infer PIIs from these texts without sophisticated reasoning.}
\label{fig:hardness-distribution}
\end{minipage}
\end{figure*}

Figures~\ref{fig:certainty-distribution} and \ref{fig:hardness-distribution} show the distribution of Certainty and Hardness levels for all PIIs, respectively. Certainty measures the confidence level for inferring PII from text on a 5-point scale from 1 (very uncertain) to 5 (very certain). Hardness measures the cognitive difficulty required to infer the PII on a 5-point scale from 1 (very easy) to 5 (very difficult).

\begin{figure*}[t]
\centering
\begin{minipage}{0.49\textwidth}
\centering
\includegraphics[width=\linewidth]{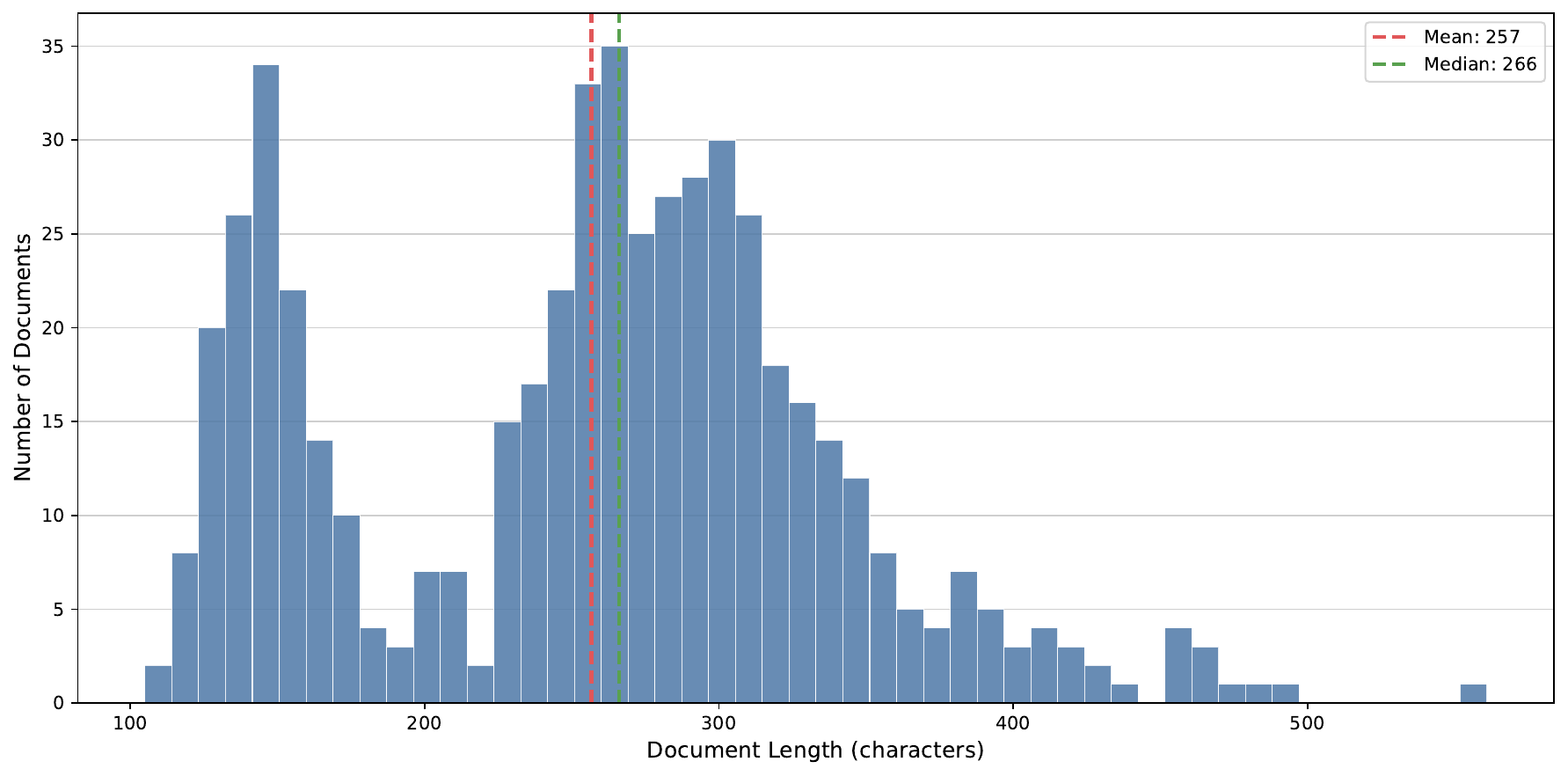}
\\[0.3em]
(a) PANORAMA
\end{minipage}
\hfill
\begin{minipage}{0.49\textwidth}
\centering
\includegraphics[width=\linewidth]{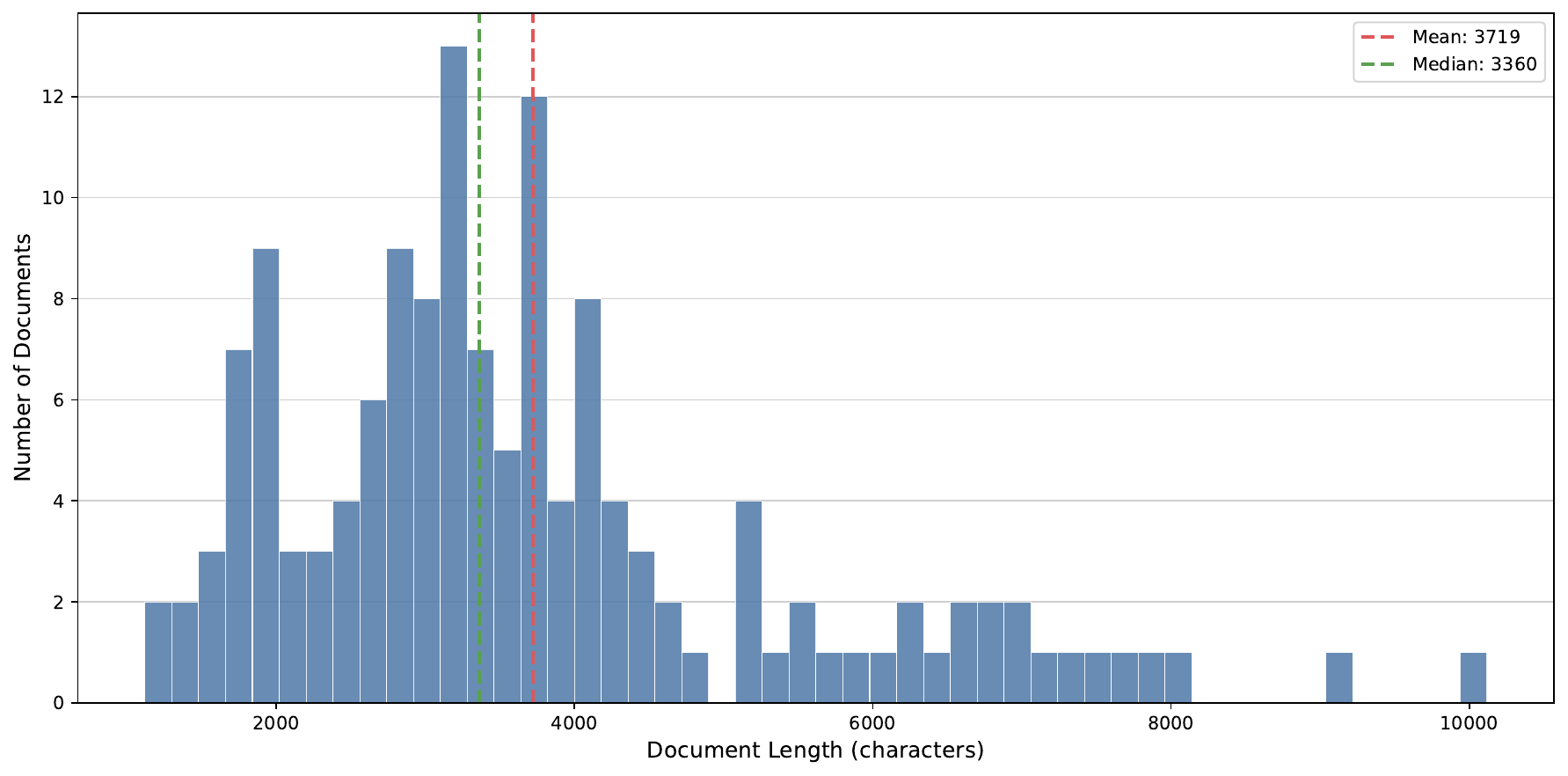}
\\[0.3em]
(b) TAB
\end{minipage}
\caption{Document length distribution. For visualization clarity, a few outliers are excluded: one PANORAMA document exceeding 1,500 characters and two TAB documents exceeding 15,000 characters.}
\label{fig:doc-length}
\end{figure*}

\subsection{Document Length Distribution}

Figure~\ref{fig:doc-length} shows the document length distribution by dataset. PANORAMA consists of short online content averaging 260 characters (99.6\% within 500 characters), while TAB consists of long legal documents averaging 3,918 characters (64.6\% in the 2,000--5,000 character range). This difference reflects the inherent characteristics of the two domains.

\subsection{Qualitative Examples}

This section presents qualitative examples for each Hardness level and Certainty level in the \ours~dataset. Hardness scores range from 1 (very easy) to 5 (very hard), while Certainty scores range from 1 (very uncertain) to 5 (very certain). For Hardness levels 4-5, annotators are permitted to use traditional online search engines. See Appendix~\ref{sec:appendix-f} for the complete grading criteria.

\noindent\textbf{Hardness Level Examples}

\begin{itemize}[leftmargin=*]
\item \textbf{Level 1}: ``The case originated in an application against the Kingdom of Sweden lodged with the Court by a Swedish national, Mr Raja Arlewin (`the applicant'), on 18 March 2010.'' $\rightarrow$ \textbf{Name: Raja Arlewin}
\item \textbf{Level 2}: ``The applicant is born in 1970 and lives in Stockholm.'' $\rightarrow$ \textbf{Age: 54-55} (calculated from birth year 1970 relative to reference date September 1, 2025)
\item \textbf{Level 3}: ``Username: Ridhi, Location: Sambalpur... Stopped by this modest cafe near my workplace.'' $\rightarrow$ \textbf{Nationality: India} (requires recognizing ``Ridhi'' as an Indian-origin name and combining it with the location context)
\item \textbf{Level 4}: ``The President of the Fourth Section decided to communicate the application to the Government.'' $\rightarrow$ \textbf{Occupation: Judge} (requires knowledge that the President of the Fourth Section at ECHR is a judge)
\item \textbf{Level 5}: ``The applicant is represented by Mr Mahmut Akdo\u{g}an, a lawyer practising in Mersin.'' $\rightarrow$ \textbf{Education: College Degree} (requires searching Turkish bar admission requirements and synthesizing that lawyers must hold a law degree)
\end{itemize}

\noindent\textbf{Certainty Level Examples}

\begin{itemize}[leftmargin=*]
\item \textbf{Level 5}: ``...a Swedish national, Mr Raja Arlewin (`the applicant').'' $\rightarrow$ \textbf{Name: Raja Arlewin} (the name is explicitly stated with no ambiguity)
\item \textbf{Level 4}: ``He is self-employed and runs a business.'' $\rightarrow$ \textbf{Occupation: Business owner} (strongly implied, though the specific business type is not stated)
\item \textbf{Level 3}: ``...Mr J. S{\"o}dergren, a lawyer practising in Stockholm.'' $\rightarrow$ \textbf{Nationality: Sweden} (practices in Stockholm suggests Swedish nationality, but foreign lawyers can also practice there)
\item \textbf{Level 2}: ``...with a job, a family and a fixed abode.'' $\rightarrow$ \textbf{Relationship: Married} (having a family suggests marriage, but could refer to being a single parent)
\item \textbf{Level 1}: ``...the applicant and Eren Keskin.'' $\rightarrow$ \textbf{Sex: Female} (the name ``Eren'' is gender-ambiguous in Turkish; without an honorific, inference is highly uncertain)
\end{itemize}

\begin{table}[t]
\centering
\small
\resizebox{\columnwidth}{!}{%
\begin{tabular}{lccc}
\toprule
\textbf{Metric} & \textbf{TAB} & \textbf{PANORAMA} & \textbf{Total} \\
\midrule
Documents & 144 & 151 & 295 \\
Num. of Subjects & 586 & 360 & 946 \\
Avg Subjects/Doc & 4.07 & 2.38 & 3.21 \\
\shortstack[l]{Num. of PIIs\\[1.65ex]~} & \shortstack[c]{3,350\\(3,064)} & \shortstack[c]{1,201\\(943)} & \shortstack[c]{4,551\\(4,007)} \\
Avg PIIs/Subject & 5.72 & 3.34 & 4.81 \\
Avg Doc Length (chars) & 3,918 & 265 & - \\
\bottomrule
\end{tabular}%
}
\caption{\ours~Test Set Basic Statistics. Numbers in parentheses indicate PIIs with Certainty $\geq$ 3.}
\label{tab:test-set-stats}
\end{table}

\begin{table*}[t]
\centering
\footnotesize
\setlength{\tabcolsep}{10pt}
\begin{minipage}{0.48\textwidth}
\centering
\begin{tabular}{lcc}
\toprule
\textbf{Model} & \textbf{Subject Match} & \textbf{PII Acc.} \\
\midrule
Claude-Sonnet-4.5 & \textbf{96.76\%} & \textbf{91.12\%} \\
GPT-4.1 & 96.42\% & 90.11\% \\
GPT-OSS-120B & 94.20\% & 88.82\% \\
Claude-Haiku-4.5 & 90.27\% & 86.05\% \\
GPT-4.1-Mini & 90.61\% & 83.09\% \\
Llama-3.1-70B & 88.91\% & 66.83\% \\
Gemma-3-27B & 86.35\% & 74.31\% \\
GPT-OSS-20B & 84.30\% & 70.92\% \\
Qwen-3-14B & 76.28\% & 68.21\% \\
Llama-3.1-8B & 76.28\% & 60.95\% \\
Gemma-3-4B & 67.92\% & 51.31\% \\
\bottomrule
\end{tabular}
\\[0.5em]
(a) TAB
\end{minipage}
\hfill
\begin{minipage}{0.48\textwidth}
\centering
\begin{tabular}{lcc}
\toprule
\textbf{Model} & \textbf{Subject Match} & \textbf{PII Acc.} \\
\midrule
Claude-Sonnet-4.5 & \textbf{96.66\%} & \textbf{91.62\%} \\
GPT-OSS-120B & 97.50\% & 82.27\% \\
GPT-4.1 & 95.47\% & 84.00\% \\
Gemma-3-27B & 95.28\% & 74.58\% \\
GPT-4.1-Mini & 92.50\% & 76.08\% \\
GPT-OSS-20B & 91.39\% & 72.88\% \\
Claude-Haiku-4.5 & 91.11\% & 85.86\% \\
Qwen-3-14B & 85.00\% & 63.68\% \\
Llama-3.1-70B & 78.33\% & 64.54\% \\
Llama-3.1-8B & 70.28\% & 68.40\% \\
Gemma-3-4B & 61.11\% & 56.21\% \\
\bottomrule
\end{tabular}
\\[0.5em]
(b) PANORAMA
\end{minipage}
\caption{Subject-level Inference Performance. Subject Match=Subject Match Ratio, PII Acc.=PII Inference Accuracy.}
\label{tab:model-comparison}
\end{table*}

\begin{figure*}[t]
\centering
\includegraphics[width=0.85\textwidth]{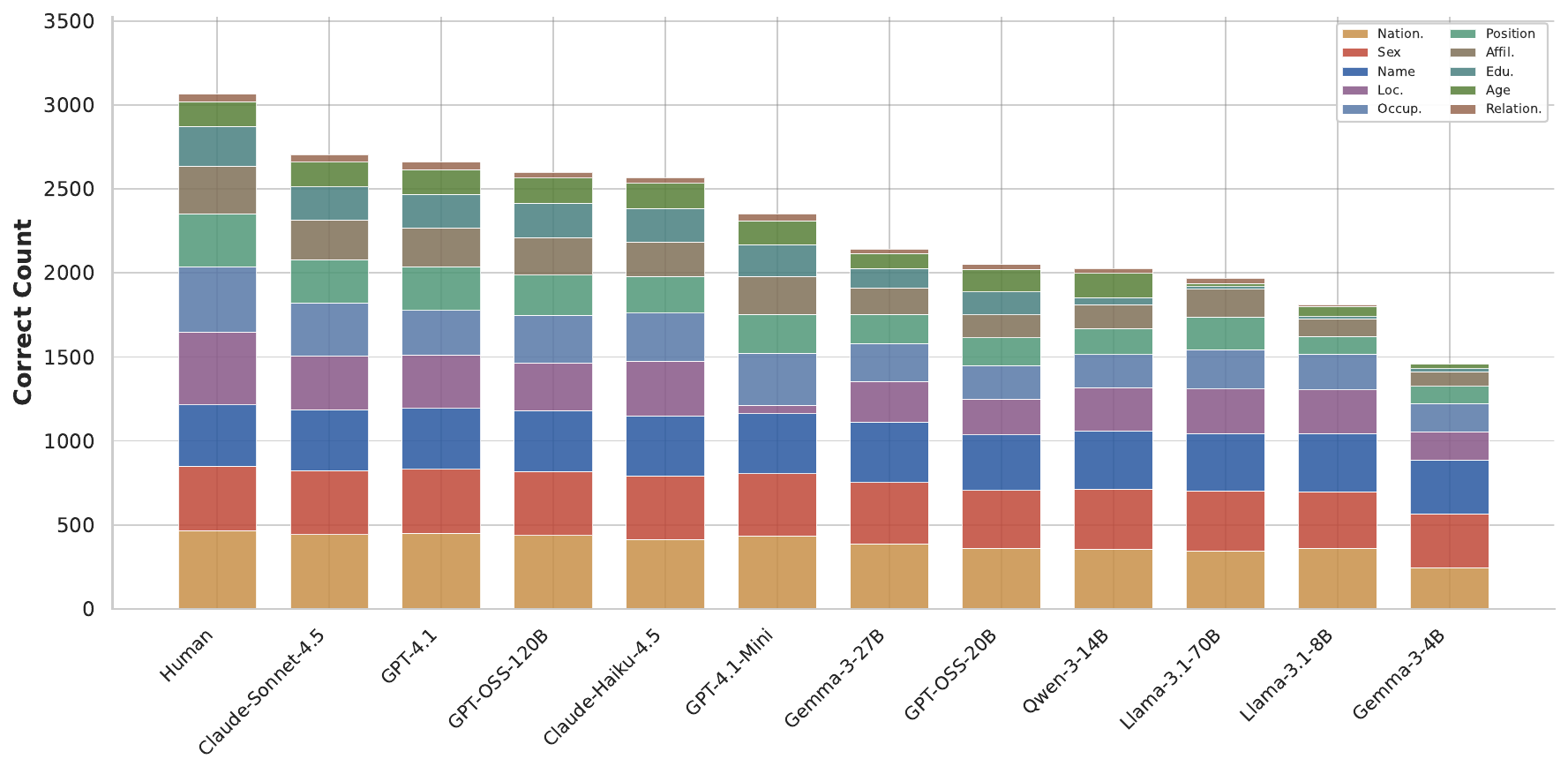}
\\[0.3em]
(a) TAB
\\[1em]
\includegraphics[width=0.85\textwidth]{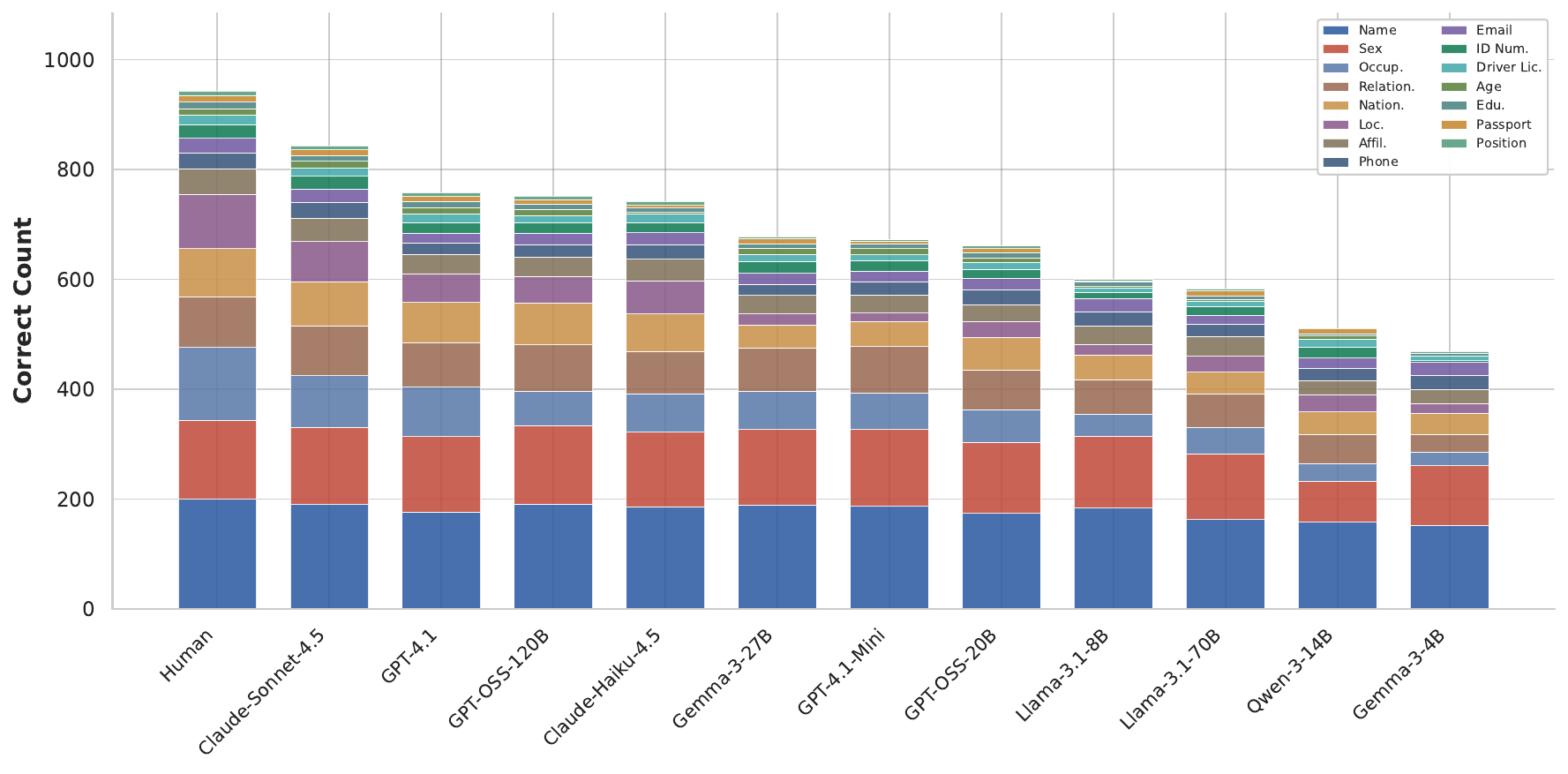}
\\[0.3em]
(b) PANORAMA
\caption{Per-tag inference accuracy counts. Each stacked bar shows the number of correctly inferred PIIs by category.}
\label{fig:inference-count}
\end{figure*}

\begin{figure*}[!t]
\centering
\includegraphics[width=0.85\textwidth]{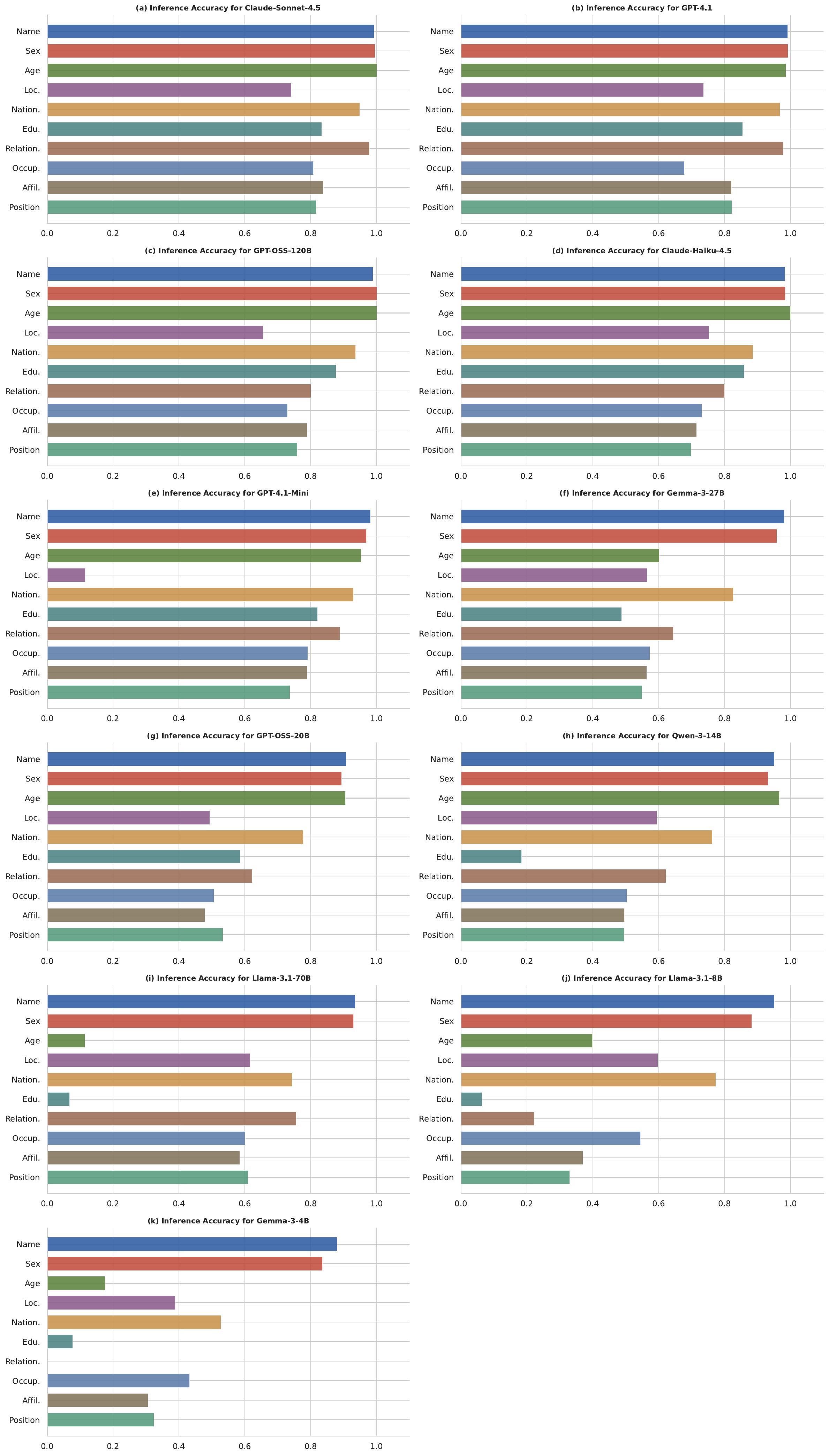}
\caption{Inference accuracy by PII category on TAB dataset.}
\label{fig:tag-accuracy-tab}
\end{figure*}

\begin{figure*}[!t]
\centering
\includegraphics[width=0.85\textwidth]{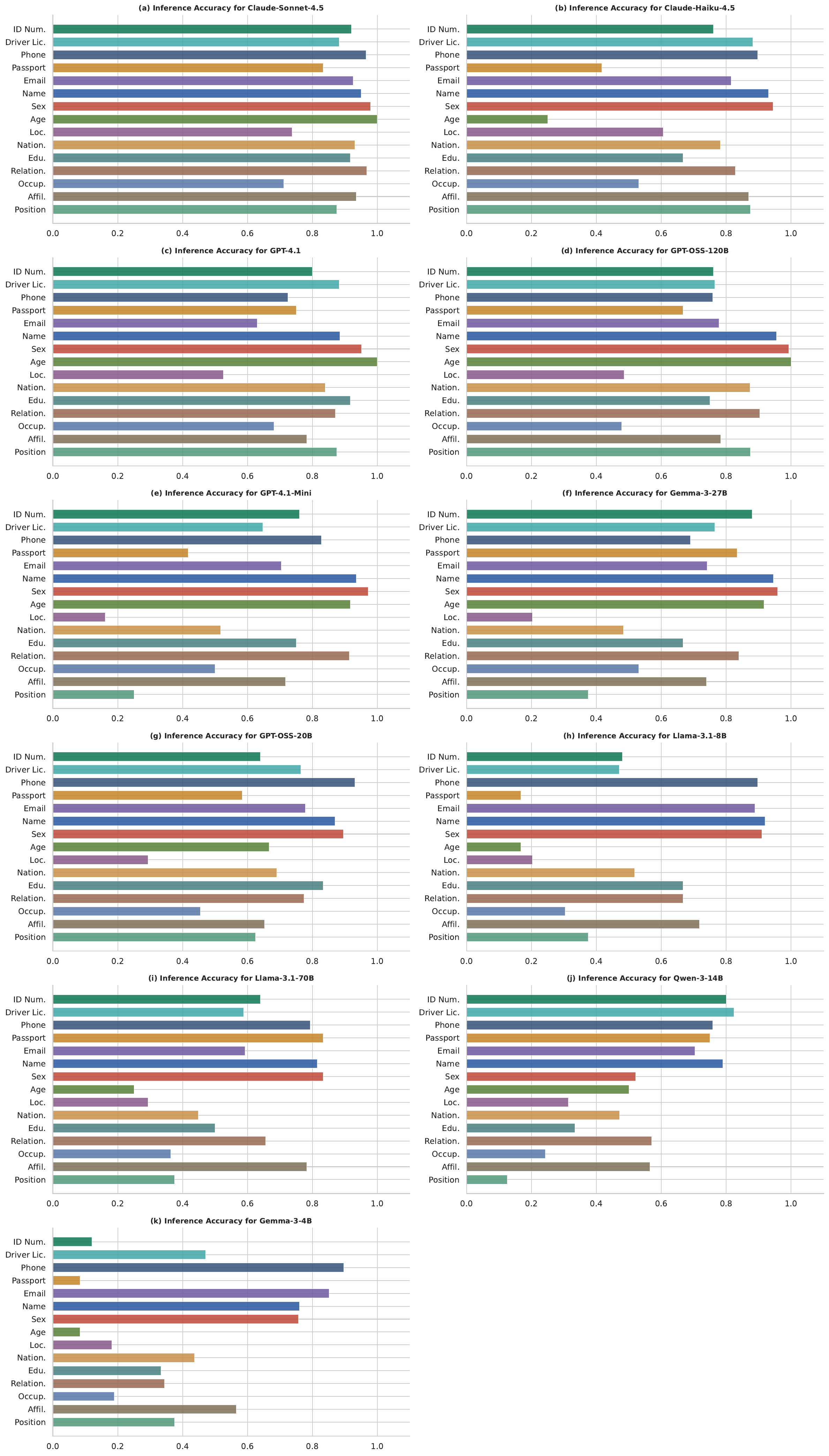}
\caption{Inference accuracy by PII category on PANORAMA dataset.}
\label{fig:tag-accuracy-panorama}
\end{figure*}

\section{Details of Subject-wise Inference Validation}
\label{sec:appendix-b}

This appendix presents detailed experimental settings and per-model comparison results for the subject-wise inference framework validation described in Section~\ref{sec:annotation-process}.

\subsection{Experimental Setup}

\noindent\textbf{Evaluation Dataset.} Manually constructed test sets (TAB 144, PANORAMA 151) are used; detailed statistics are provided in Table~\ref{tab:test-set-stats}. Consistent with \citet{staab2024beyond,staab2025anonymizers}, only labels with Certainty $\geq$ 3 are selected for evaluation.

\noindent\textbf{Inference LLM.} We evaluate 11 LLMs at temperature 0.1~\citep{staab2024beyond,staab2025anonymizers} to select the inference LLM for subject-wise PII extraction:
\begin{itemize}[leftmargin=*]
\item \textbf{Proprietary}: Claude Sonnet 4.5 (\texttt{claude-sonnet-4-5-20250929}) and Claude Haiku 4.5 (\texttt{claude-haiku-4-5-20251001}) \citep{anthropic2025claude}; GPT-4.1 (\texttt{gpt-4.1-2025-04-14}) and GPT-4.1 mini (\texttt{gpt-4.1-mini-2025-04-14}) \citep{openai2025gpt41}.
\item \textbf{Open-source}: openai/gpt-oss-120b and openai/gpt-oss-20b \citep{openai2025gptoss}; google/gemma-3-27b-it and google/gemma-3-4b-it \citep{gemmateam2025gemma3}; meta-llama/Llama-3.1-70B-Instruct and meta-llama/Llama-3.1-8B-Instruct \citep{grattafiori2024llama3}; Qwen/Qwen3-14B \citep{qwen2025qwen3}.
\end{itemize}

\noindent\textbf{Evaluator LLM.} GPT-4.1-Mini serves as the evaluator LLM for subject alignment and PII comparison, offering favorable cost and processing speed.

\noindent\textbf{Evaluation Metrics.}
\begin{itemize}[leftmargin=*]
\item \textbf{Subject Match Ratio}: Matching success rate between ground truth subjects and LLM-inferred subjects
\item \textbf{Inference Accuracy}: PII inference accuracy for matched subjects (Top-1 prediction)
\end{itemize}

\subsection{Model Performance Overview}

Table~\ref{tab:model-comparison} shows the subject-level inference performance of 11 models on TAB and PANORAMA datasets. Claude-Sonnet-4.5 achieves the highest performance on both datasets and is therefore selected as the pre-labeling and anonymization evaluation model.

\noindent\textbf{Model Scale and Performance.} Overall, larger models show higher performance, and proprietary models outperform open-source models. Claude-Sonnet-4.5 achieves the highest performance on both datasets and is selected as the inference model for subsequent pre-labeling and anonymization evaluation.

\noindent\textbf{Dataset Characteristics.} Both datasets show comparable overall performance. TAB achieves slightly higher average Inference Accuracy (75.6\% vs 74.6\%), while PANORAMA shows slightly higher average Subject Match Ratio (86.8\% vs 86.2\%), reflecting the different characteristics of legal documents versus social media texts.

\noindent\textbf{Subject Identification Capability.} Subject Match Ratio is higher than Inference Accuracy across all models, suggesting that subject identification is relatively easier than PII inference. However, for smaller models (8B and below), Subject Match Ratio also drops below 85\%, indicating that sufficiently large models are required for accurate subject-level evaluation.

\subsection{Analysis by PII Category}

Figure~\ref{fig:inference-count} shows the per-tag breakdown of correctly inferred PIIs for each model. Human ground truth represents the total PII counts in the ground truth, while other bars show the number of PIIs each model correctly inferred. Figures~\ref{fig:tag-accuracy-tab} and \ref{fig:tag-accuracy-panorama} show the inference accuracy by PII category for each model.

\subsection{Analysis by Hardness Level}

\begin{figure*}[t]
\centering
\includegraphics[width=0.85\textwidth]{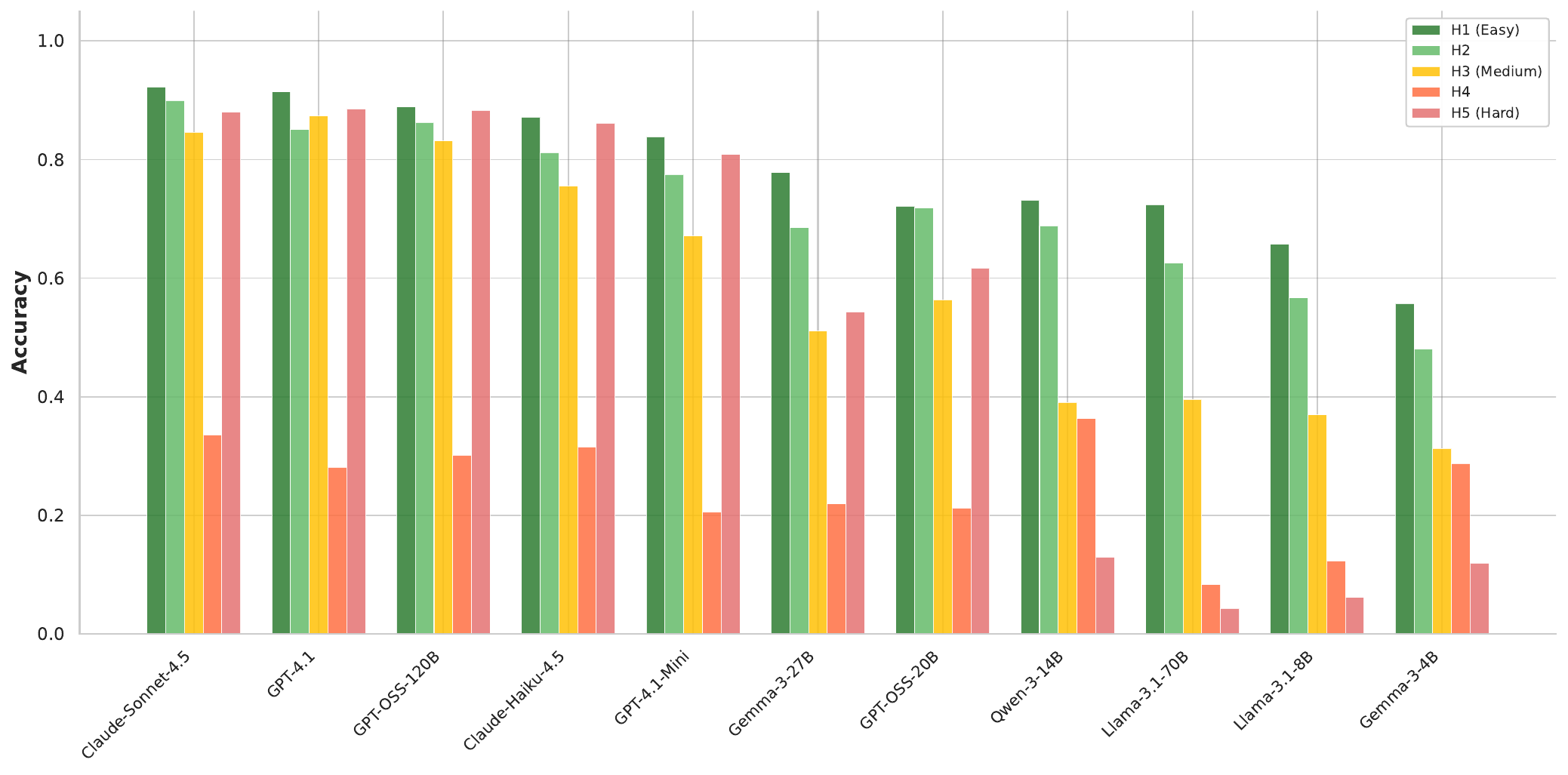}
\\[0.3em]
(a) TAB
\\[1em]
\includegraphics[width=0.85\textwidth]{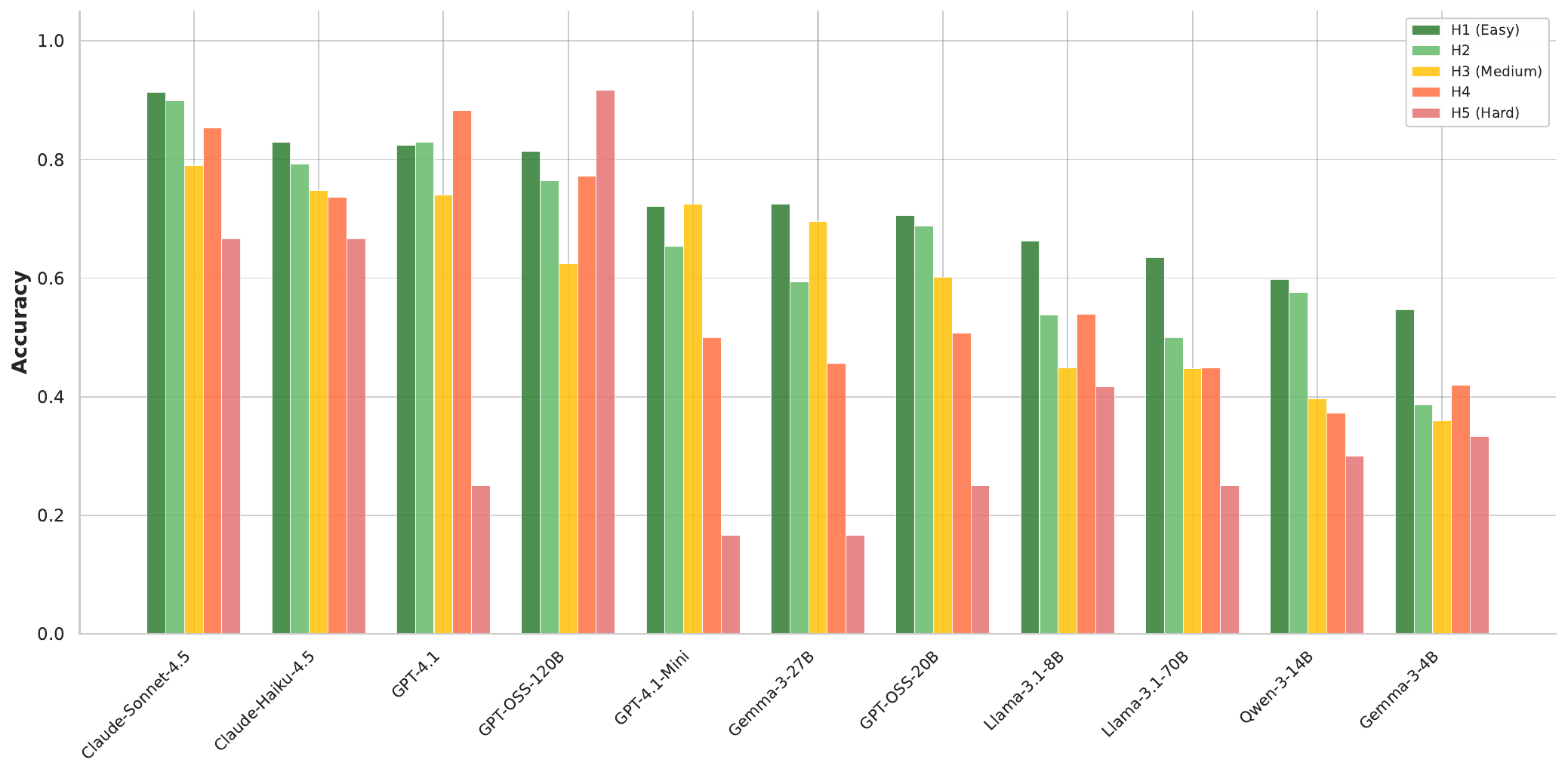}
\\[0.3em]
(b) PANORAMA
\caption{Inference accuracy by Hardness level.}
\label{fig:hardness-accuracy}
\end{figure*}

Figure~\ref{fig:hardness-accuracy} shows the inference accuracy by Hardness level for TAB and PANORAMA datasets. A decreasing trend in inference accuracy is observed as Hardness increases, suggesting that LLM inference performance degrades for cognitively more difficult PIIs.

\section{Experiment Details}
\label{sec:appendix-c}

This appendix describes the detailed experimental settings and evaluation methodology used in Sections~\ref{sec:eval-framework} and \ref{sec:experiments}.

\subsection{Test Set Statistics}
\label{sec:appendix-c1}

The test set from the \ours~benchmark described in Section~\ref{sec:benchmark} is used for evaluation. Table~\ref{tab:test-set-stats} shows the basic statistics of the test set.

\subsection{Hardware and Software}

Experiments were conducted on Intel Xeon Gold 6448Y, 221GB DDR5, NVIDIA H100 80GB GPU environment. Python 3.11+ and CUDA 12.2 were used.

\subsection{Subject-level Inference Evaluation}
\label{sec:appendix-c3}

\begin{figure*}[t]
\centering
\includegraphics[width=0.7\textwidth]{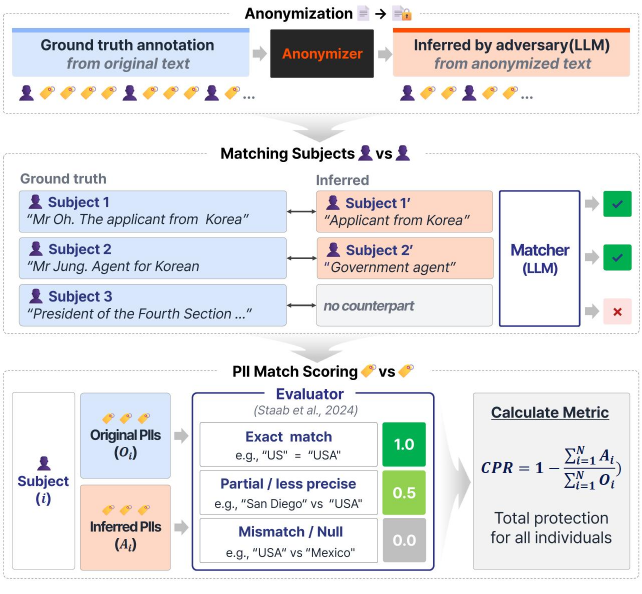}
\caption{Overview of the subject-level inference evaluation protocol, illustrated with examples.}
\label{fig:eval-protocol}
\end{figure*}

This section describes the detailed implementation of the 3-step evaluation pipeline defined in Section~\ref{sec:eval-protocol}. Based on the validation results in Appendix~\ref{sec:appendix-b}, Claude-Sonnet-4.5 is selected as the adversarial LLM, achieving the highest Inference Accuracy (above 91\%) and Subject Match Ratio (above 96\%) on both datasets. GPT-4.1-Mini serves as the evaluator LLM, as used in framework validation (Appendix~\ref{sec:appendix-b}). Following prior work~\citep{staab2024beyond,staab2025anonymizers}, only PIIs with Certainty $\geq$ 3 are evaluated, using the top-1 inference (the model's single best guess) for comparison. Figure~\ref{fig:eval-protocol} provides an overview of this protocol with illustrative examples, showing how the adversary's inference on anonymized text is processed through subject matching and per-PII scoring to compute the final protection metric (CPR).

\subsubsection{Subject-level Comparison}

The adversarial LLM extracts subjects and PIIs from anonymized text using the two-stage framework (Section~\ref{sec:annotation-process}): subject identification followed by PII inference for 15 categories. The prompts are shown in Figures~\ref{fig:prompt-subject-id}--\ref{fig:prompt-noncode-pii}.

One-to-one correspondence between Ground Truth subjects (annotated on original text) and subjects identified from anonymized text is then established by the evaluator LLM based on subject descriptions and the text content. Anonymization can remove explicit identifiers, so matching considers roles and contextual cues in addition to explicit mentions. Ground Truth subjects that fail to match are assigned 0.0 points for all PIIs. The alignment prompt for matching subjects is shown in Figure~\ref{fig:prompt-subject-align-anon}.

The following example illustrates subject matching between Ground Truth and inference from anonymized text:
\begin{itemize}[leftmargin=*]
\item \textbf{Subject 0}: Ground Truth describes ``Mr Jan Kowalski - The applicant, Polish national born in 1934, lives in Warsaw, ...'' $\rightarrow$ Matched with ``The applicant - Individual who lodged the application ...''
\item \textbf{Subject 1}: Ground Truth describes ``Mr Stefan Nowak - Agent representing the Polish Government, ...'' $\rightarrow$ Matched with ``Agent representing the Government - Specific official designated to represent ...''
\end{itemize}

For framework validation (Appendix~\ref{sec:appendix-b}), inference is performed on original text, and subject matching uses the alignment prompt in Figure~\ref{fig:prompt-subject-align-same}. Human annotators verified the matching results to calculate Subject Match Ratio and validate the matching procedure.

\subsubsection{PII-level Comparison}

For matched subject pairs, PII-level comparison follows \citet{staab2024beyond}'s scoring scheme:
\begin{itemize}[leftmargin=*]
\item Rule-based comparison is first performed by category: free text (Name, Occupation, etc.) uses Jaro-Winkler similarity (threshold 0.85), Age uses $\pm$5 year range matching, Location uses hierarchical matching, and CODE types use exact matching after normalization.
\item Cases judged as mismatch (0.0 points) are re-evaluated for semantic equivalence using the evaluator LLM, which assigns Match (1.0), Less Precise (0.5), or Mismatch (0.0). The scoring prompt is shown in Figure~\ref{fig:prompt-pii-agree-gt}.
\item Human evaluation is triggered only when the LLM evaluator fails to return a valid response due to response errors or parsing exceptions; for framework validation, all mismatch cases were manually reviewed and evaluated by human annotators.
\end{itemize}

\subsubsection{CPR and IPR Calculation Example}
\label{sec:cpr-ipr-example}

Consider a document with 3 Ground Truth subjects containing a total of 9 PIIs. After anonymization, the adversarial LLM attempts to identify subjects and infer their PIIs. Subject matching is performed by comparing Ground Truth subject descriptions with LLM-inferred subject descriptions via the alignment prompt (Figure~\ref{fig:prompt-subject-align-anon}). Each inferred PII is then scored as 1.0 (exact match), 0.5 (less precise), or 0.0 (mismatch or not inferred) via the PII agreement evaluation prompt (Figure~\ref{fig:prompt-pii-agree-gt}):

\begin{itemize}[leftmargin=*]
\item \textbf{Subject 1 (matched):} Originally had 4 PIIs ($O_1 = 4$). The adversary matched this subject and inferred 4 PIIs with scores: 1.0 (exact match), 0.5 (less precise), 0.5 (less precise), 0.0 (mismatch). Total inferred $A_1 = 2.0$.

\item \textbf{Subject 2 (matched):} Originally had 2 PIIs ($O_2 = 2$). The adversary matched this subject and inferred 2 PIIs with scores: 1.0 (exact match), 0.5 (less precise). Total inferred $A_2 = 1.5$.

\item \textbf{Subject 3 (unmatched):} Originally had 3 PIIs ($O_3 = 3$). The adversary failed to identify this subject from the anonymized text. Per Step 1, all PIIs are assigned 0.0. Total inferred $A_3 = 0.0$.
\end{itemize}

\noindent\textbf{CPR} measures the proportion of protected PIIs across all subjects. The adversary inferred a total of $A_1 + A_2 + A_3 = 2.0 + 1.5 + 0.0 = 3.5$ PIIs out of $O_1 + O_2 + O_3 = 4 + 2 + 3 = 9$ Ground Truth PIIs. Thus, $\text{CPR} = 1 - 3.5/9 \approx 0.611$.

\noindent\textbf{IPR} averages per-subject protection rates equally. Each subject's protection rate is: Subject 1: $1 - 2.0/4 = 0.50$, Subject 2: $1 - 1.5/2 = 0.25$, Subject 3: $1 - 0.0/3 = 1.00$. Thus, $\text{IPR} = (0.50 + 0.25 + 1.00)/3 \approx 0.583$.

The unmatched subject contributes 1.0 to IPR, reflecting complete protection when the adversary cannot identify the subject's existence. In this example, CPR (0.611) exceeds IPR (0.583) because IPR assigns equal weight to each subject, making it more sensitive to underprotected subjects like Subject 2, whereas CPR weights proportionally to PII count.

\subsection{Token and Entity Recall Evaluation}

The TAB benchmark \citep{pilan2022tab} evaluation methodology is applied to measure token and entity-level masking performance.

\noindent\textbf{Ground Truth Construction.} For Token Recall and Entity Recall calculation, entity ground truth based on the TAB 8-category system (PERSON, CODE, LOC, ORG, DEM, DATETIME, QUANTITY, MISC) is used. The TAB dataset uses the original benchmark's entity annotations as-is. Since the PANORAMA dataset does not have entity annotations, separate entity annotation is performed in this study following the TAB benchmark guidelines; the annotation procedure and quality verification results are described in Appendix~\ref{sec:appendix-d}.

\noindent\textbf{Category System Compatibility.} The anonymization experiments target TAB's 8 categories, while inference evaluation targets SPIA's 15 categories. These systems are compatible because TAB's categories encompass SPIA's: TAB's DEM category covers demographic attributes (age, gender, occupation, education), and MISC covers all other personal information. As \citet{pilan2022tab} defined annotation targets as ``all text elements with re-identification risk,'' text anonymized under TAB's 8 categories should protect information corresponding to SPIA's 15 categories.

\noindent\textbf{Masking Detection.} For TAB Longformer, entity information is preserved during masking and is used directly; for DeID-GPT, DP-Prompt, and AA, we use span search—checking whether each ground-truth entity's text still exists in the anonymized output to determine masking.

\subsection{Utility Evaluation}

\citet{staab2024beyond}'s utility evaluation methodology is applied to measure the degree of meaning preservation in anonymized text.

\noindent\textbf{Metrics.} Following \citet{staab2024beyond}, we use LLM-based Readability (1--10) and Meaning preservation (1--10) scores, along with ROUGE-L (longest common subsequence F1). The integrated utility metric is calculated as $\text{Mean Utility} = (\text{Readability} + \text{Meaning} + \text{ROUGE-L}) / 3$, where LLM scores are normalized to [0,1].

\subsection{Single-subject Evaluation (1-AAC)}

For comparison with prior work \citep{staab2024beyond}, the single-subject evaluation metric 1-AAC (1 - Adversarial Accuracy) is also reported.

\noindent\textbf{Target Subject.} The target subject is defined as the applicant for TAB and the author for PANORAMA. For 1-AAC calculation, single-subject ground truth data is separately constructed by extracting only the target subject for each dataset from the original per-subject ground truth data.

\noindent\textbf{Calculation.} Calculated as $\text{1-AAC} = 1 - \sum S / \sum A$, where $S$ is the sum of PII scores successfully inferred from anonymized text and $A$ is the total number of Ground Truth PIIs from original text.

\subsection{Method Implementation Details}

Four anonymization techniques are compared and evaluated in this study.

\noindent\textbf{TAB Longformer.} The Longformer-based NER model from \citet{pilan2022tab} is used. Based on \texttt{allenai/longformer-base-4096} \citep{beltagy2020longformer}, a confidence threshold of 0.55 is applied following the original paper's settings. For fair evaluation, the model is retrained excluding the TAB test set (N=144) used in this experiment from the training data.

\noindent\textbf{DeID-GPT.} The zero-shot prompting-based anonymization technique from \citet{liu2023deidgpt} is applied. The original paper's 18 PII categories are restructured to the TAB 8-category system, and temperature 0.05 is used to induce deterministic output following the original paper's settings. The prompt is described in Appendix~\ref{sec:appendix-g11}.

\noindent\textbf{DP-Prompt.} The differential privacy-based paraphrasing approach from \citet{utpala2023locally} is applied. High temperature (1.5, or 1.0 due to Anthropic API limitations) and top\_p 1.0 are used for linguistic pattern obfuscation following the original paper's settings. The prompt is described in Appendix~\ref{sec:appendix-g12}.

\noindent\textbf{AA.} The feedback-guided iterative anonymization technique from \citet{staab2024beyond} is applied. Following the original paper's settings, prompt level 3 (Chain-of-Thought) is used with 3 iterative refinement rounds, and the original paper's Reddit author attributes are restructured to TAB's 8 categories. For each dataset, TAB is set with ``applicant'' as the target subject for ``legal case document,'' and PANORAMA is set with ``author'' as the target subject for ``text written by one author.'' The prompt is described in Appendix~\ref{sec:appendix-g13}.

Since the performance of anonymization methods heavily depends on the backbone used \citep{staab2024beyond}, the same set of 6 backbones---GPT-4.1, GPT-4.1-Mini, Claude-Sonnet-4.5, Claude-Haiku-4.5, Llama-3.1-8B, and Gemma-3-27B---is applied to all generative methods to isolate and evaluate the effectiveness of the methodology itself.

\section{Annotation Quality for Span-based Evaluation}
\label{sec:appendix-d}

The PANORAMA dataset does not include entity annotations in its original form. Therefore, we performed separate entity annotation for span-based metrics (Token Recall, Entity Recall) evaluation in Section~\ref{sec:experiments}.

\subsection{Annotation Procedure}

\noindent\textbf{Annotation Guidelines.} We applied the TAB benchmark \citep{pilan2022tab} annotation guidelines to annotate 8 entity categories (PERSON, CODE, LOC, ORG, DEM, DATETIME, QUANTITY, MISC) and identifier types (DIRECT, QUASI).

\noindent\textbf{Annotators.} Two experts in privacy protection participated in the annotation work.

\begin{figure*}[t]
\centering
\includegraphics[width=0.90\textwidth]{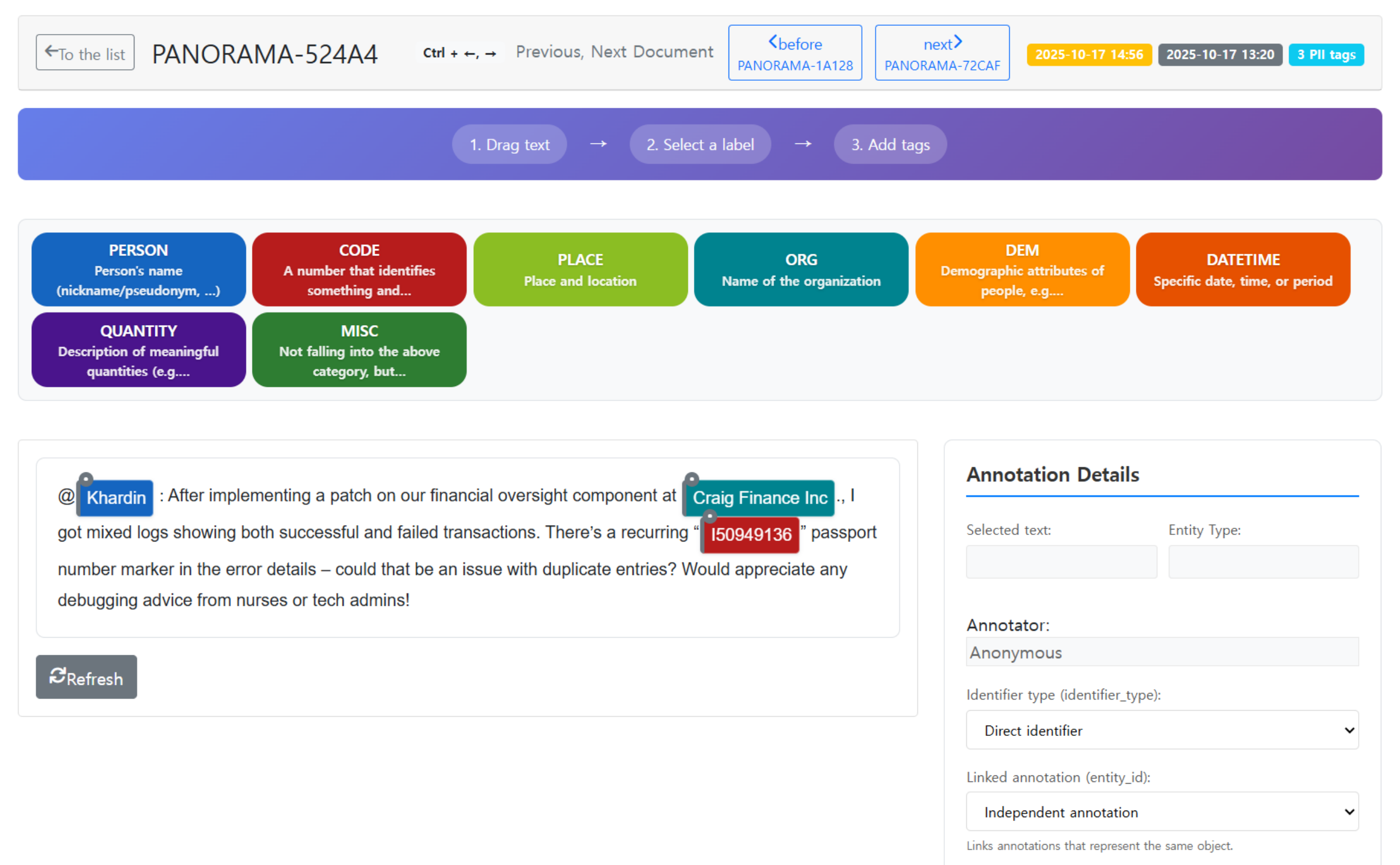}
\caption{Entity annotation tool interface. Annotators can select spans from text and specify entity type and identifier type.}
\label{fig:entity-labeler}
\end{figure*}

\noindent\textbf{Annotation Tool.} We developed a custom web-based tool for entity annotation. Figure~\ref{fig:entity-labeler} shows the tool interface.

\noindent\textbf{Annotation Procedure.} Approximately 20\% (31 documents) of the total 151 documents were assigned to both annotators for cross-labeling. The average annotation time is approximately 2 minutes per document. After annotation completion, disagreements are resolved through third-party review.

\subsection{Inter-Annotator Agreement}

To verify annotation quality, we measured inter-annotator agreement using the same span-level Average Observed Agreement (AOA) method as \citet{pilan2022tab}. Table~\ref{tab:annotation-agreement} compares the agreement results for the 31 overlapping documents (139 entities) with the TAB benchmark.

\begin{table}[t]
\centering
\small
\begin{tabular}{lcc}
\toprule
\textbf{Metric} & \textbf{PANORAMA} & \textbf{TAB} \\
\midrule
Entity Type Exact Match & 85.6\% & 75.0\% \\
Entity Type Partial Match & 86.3\% & 80.0\% \\
Identifier Type Exact Match & 79.9\% & 67.0\% \\
Identifier Type Partial Match & 81.3\% & 71.0\% \\
\bottomrule
\end{tabular}
\caption{PANORAMA Entity Annotation Agreement (vs. TAB Benchmark).}
\label{tab:annotation-agreement}
\end{table}

The higher agreement for PANORAMA compared to TAB is attributed to the relatively shorter text length (average 260 vs. 3,918 characters) and fewer entities per document, resulting in lower annotation difficulty.

\noindent\textbf{Note}: The TAB dataset uses entity labels from the original benchmark as-is.

\section{Additional Results}
\label{sec:appendix-e}

This appendix presents additional analysis results mentioned in Section~\ref{sec:experiments}.

\subsection{Multi-Adversary Robustness Analysis}
\label{sec:multi-adversary-full}

\begin{figure*}[t]
\centering
\includegraphics[width=\textwidth]{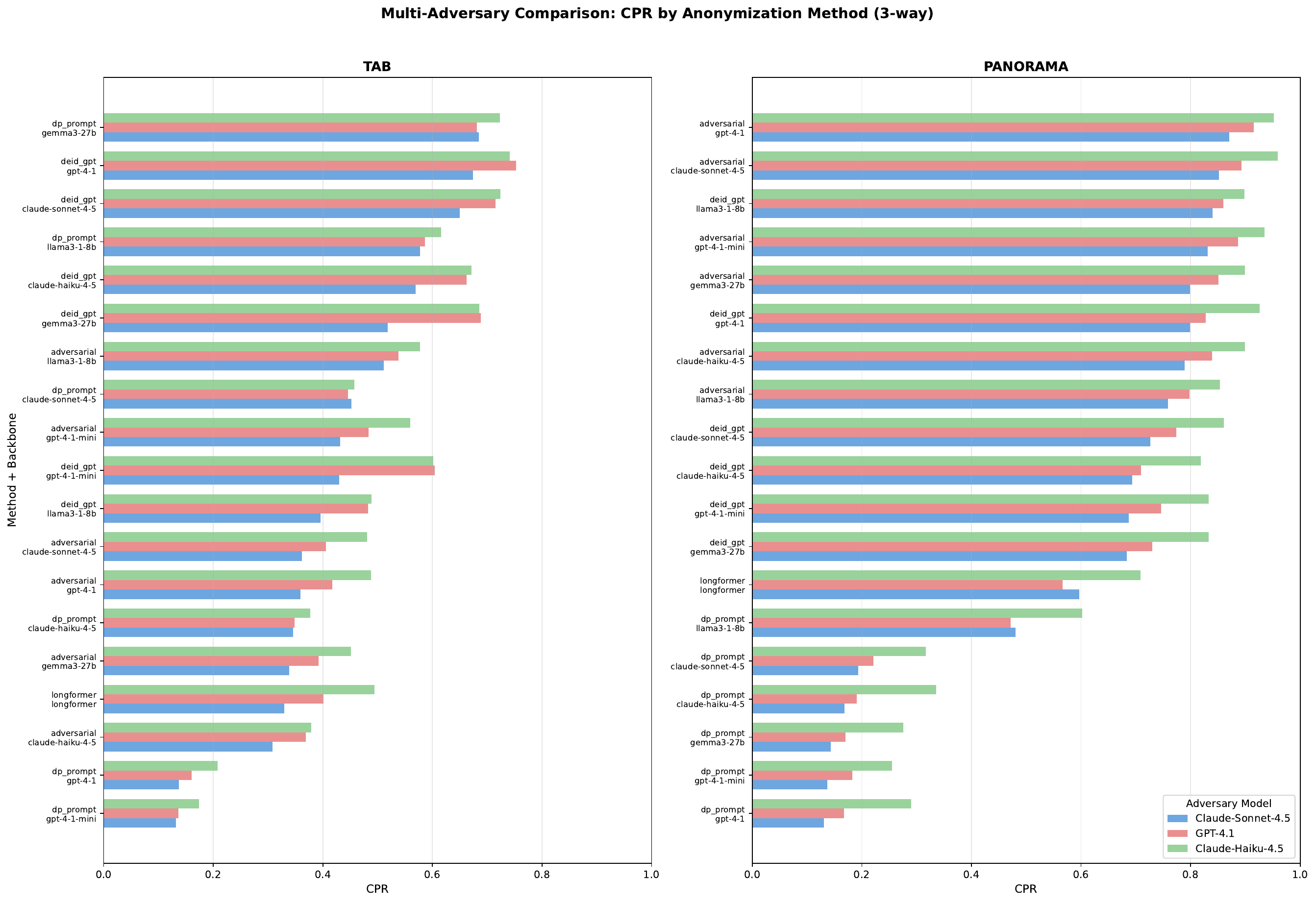}
\caption{CPR by anonymization method and backbone across three adversary models. Within each dataset, configurations are ordered by their CPR under Claude-Sonnet-4.5 (ascending).}
\label{fig:multi-adversary-cpr}
\end{figure*}

To verify that the evaluation is not biased by the choice of a single adversary, we varied the adversary across two additional models---GPT-4.1 and Claude-Haiku-4.5---alongside Claude-Sonnet-4.5. Table~\ref{tab:adversary-correlation} reports pairwise Spearman rank correlations of CPR and IPR across all anonymization configurations ($n=38$), and Figure~\ref{fig:multi-adversary-cpr} visualizes CPR for all anonymization method--backbone combinations across the three adversaries. All pairs yield $\rho > 0.98$, indicating that relative rankings among anonymization methods remain highly consistent regardless of adversary choice. The absolute CPR/IPR gaps between adversaries range from 1.3 to 4.6 percentage points for representative high-performing configurations.

\begin{table}[h]
\centering
\small
\resizebox{\columnwidth}{!}{%
\begin{tabular}{lcc}
\toprule
\textbf{Adversary Pair} & \textbf{CPR $\rho$} & \textbf{IPR $\rho$} \\
\midrule
Claude-Sonnet-4.5 vs.\ GPT-4.1 & .980 & .981 \\
Claude-Sonnet-4.5 vs.\ Claude-Haiku-4.5 & .980 & .980 \\
GPT-4.1 vs.\ Claude-Haiku-4.5 & .986 & .981 \\
\bottomrule
\end{tabular}}
\caption{Spearman rank correlation of CPR and IPR across adversary models ($n{=}38$, all $p{<}10^{-26}$).}
\label{tab:adversary-correlation}
\end{table}

\subsection{PII Type-wise Analysis (CODE vs NON-CODE)}

We analyzed protection rates by categorizing PII into CODE and NON-CODE types based on morphological characteristics. Since the TAB dataset does not contain CODE-type PIIs, we conducted analysis only on the PANORAMA dataset, which includes CODE types.

\noindent\textbf{PII Type Definitions:}
\begin{itemize}[leftmargin=*]
\item \textbf{CODE types} (5): ID Number, Driver License, Phone, Passport, Email
\item \textbf{NON-CODE types} (10): Name, Sex, Age, Location, Nationality, Education, Relationship, Occupation, Affiliation, Position
\end{itemize}

\begin{figure*}[t]
\centering
\includegraphics[width=0.75\textwidth]{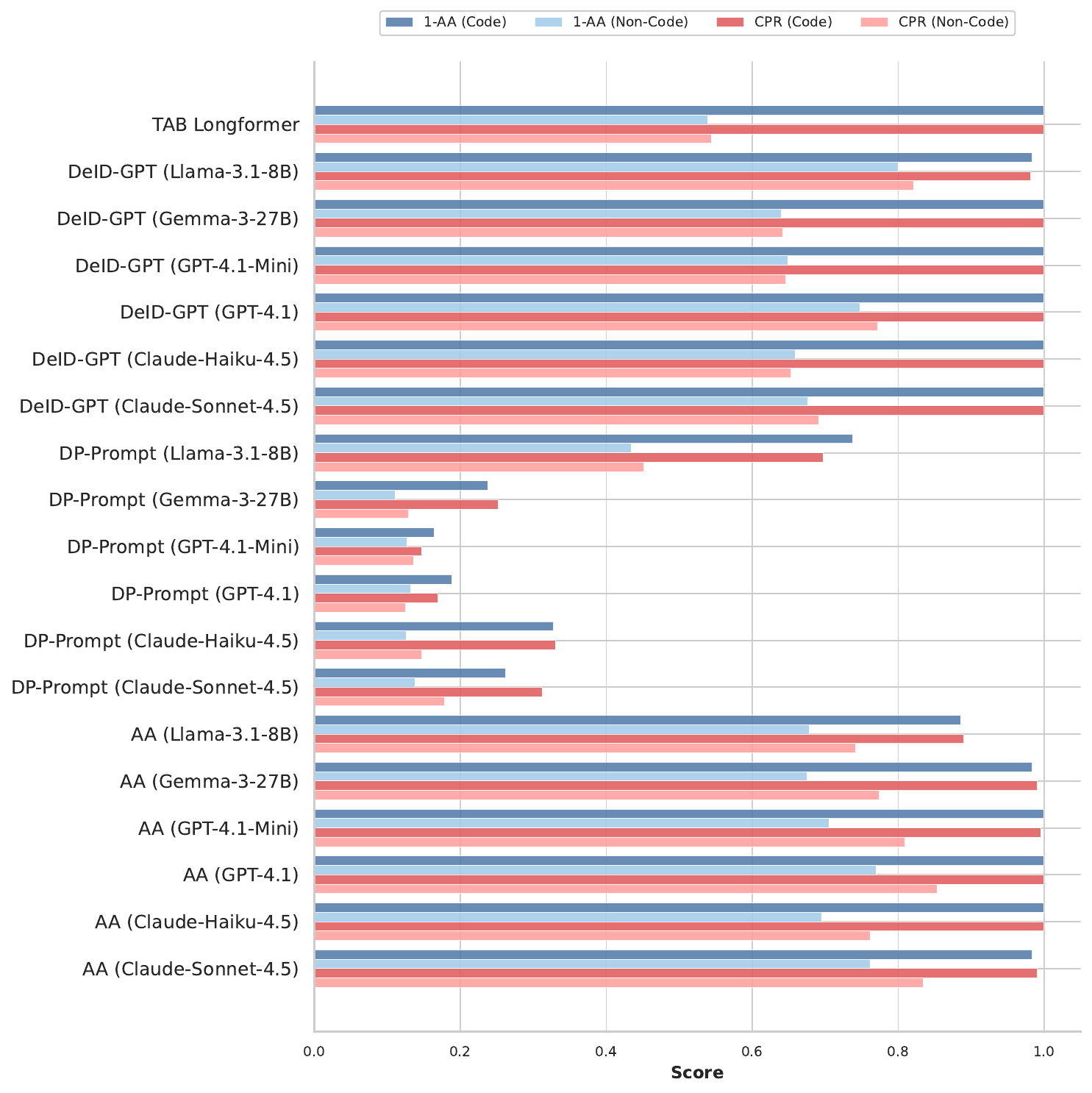}
\caption{PII type-wise protection rate analysis on PANORAMA dataset. CODE-type PIIs achieve CPR 1.0 in most techniques, while NON-CODE types show relatively lower protection rates.}
\label{fig:pii-type-analysis}
\end{figure*}

Figure~\ref{fig:pii-type-analysis} shows the protection rate analysis results by PII type. CODE-type PIIs achieve CPR 1.0 in most techniques, while NON-CODE types show relatively lower protection rates. This demonstrates that NON-CODE-type PIIs can be indirectly inferred from context and are not fully protected by span masking alone.

\subsection{Inference by Hardness Level After Anonymization}
\label{sec:hardness-anon-analysis}

\begin{figure*}[t]
\centering
\includegraphics[width=0.95\textwidth]{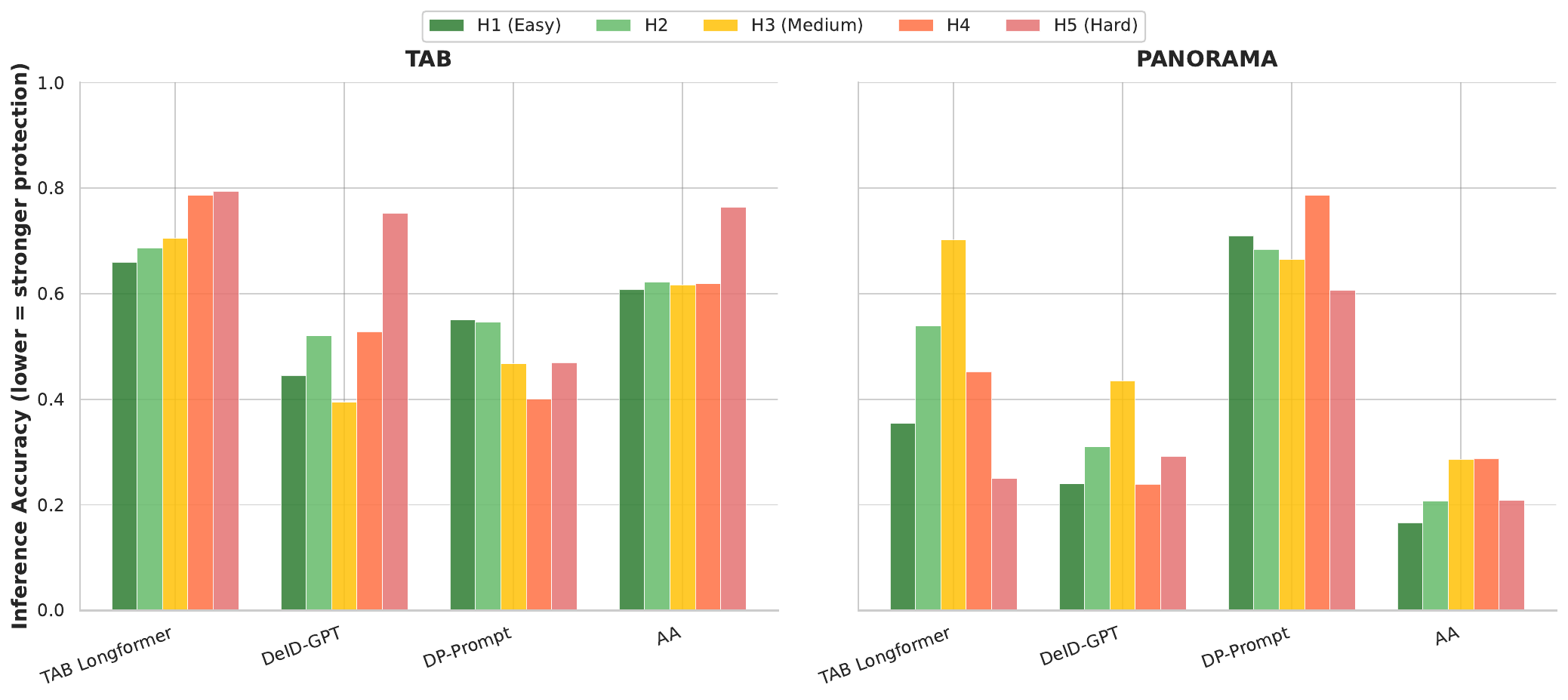}
\caption{Inference accuracy by Hardness level after anonymization,
averaged across all backbones. Lower is stronger protection. TAB and PANORAMA
diverge at Hardness levels 4--5 for three of four methods.}
\label{fig:hardness-anon}
\end{figure*}

On original text, inference accuracy shows a decreasing trend as hardness
increases (Figure~\ref{fig:hardness-accuracy}). After anonymization, this trend no longer holds uniformly across
datasets (Figure~\ref{fig:hardness-anon}): on TAB, three of four methods leave
higher-hardness PIIs (levels 4--5) \emph{as---or more---inferable} than lower-hardness
ones, whereas on PANORAMA the same methods hold or tighten protection at Hardness levels 4--5.

This divergence reflects how anonymization interacts with each domain's
document structure rather than the higher-Hardness categories alone.
Anonymizers substitute identifier-like tokens (names, profession words, place
names) with [redacted], but leave intact contextual descriptors such as
\emph{``representing the applicant''} or \emph{``Agent of the
Government''}---because these are not themselves recognized as PII. In TAB,
such phrases survive in abundance within \textasciitilde4{,}000-character legal
documents: even after \emph{``lawyer''} is redacted, nearby role descriptions
let the underlying profession---and hence the inferred education---remain
inferable. PANORAMA, by contrast, encodes higher-Hardness cues within
\textasciitilde260-character posts as explicit personal cues such as
\emph{``\#ProfessorVibes''} or \emph{``flight simulation logs''}; these are
identified and removed in a single substitution. Higher-Hardness inferability
after anonymization thus depends on how much contextual structure is preserved
during the process---substantially more in long, structurally rich legal text.

\subsection{Privacy-Utility Trade-off Analysis}

\begin{figure*}[t]
\centering
\begin{minipage}{0.48\textwidth}
\centering
\includegraphics[width=\textwidth]{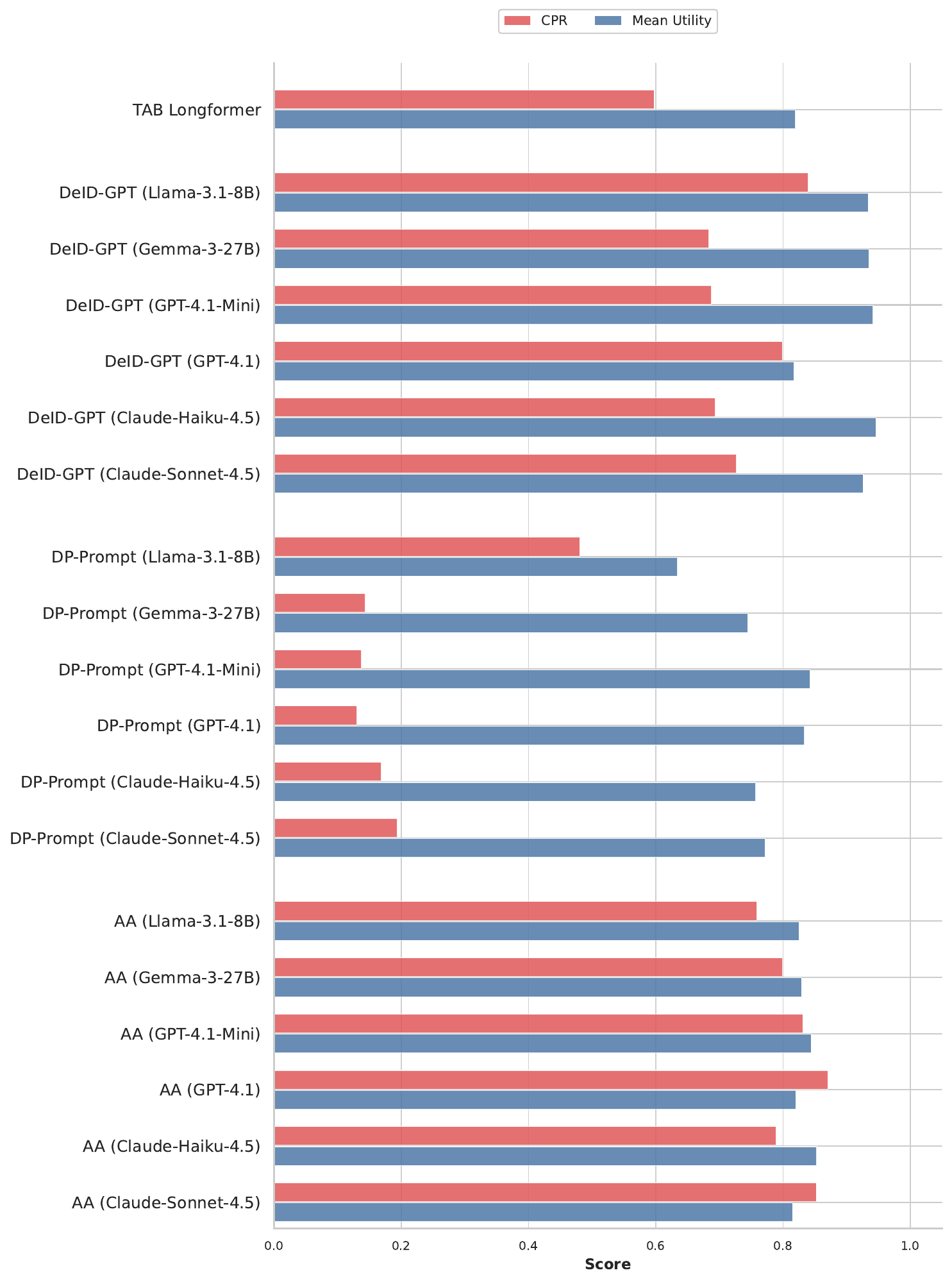}
\\[0.3em]
(a) PANORAMA
\end{minipage}
\hfill
\begin{minipage}{0.48\textwidth}
\centering
\includegraphics[width=\textwidth]{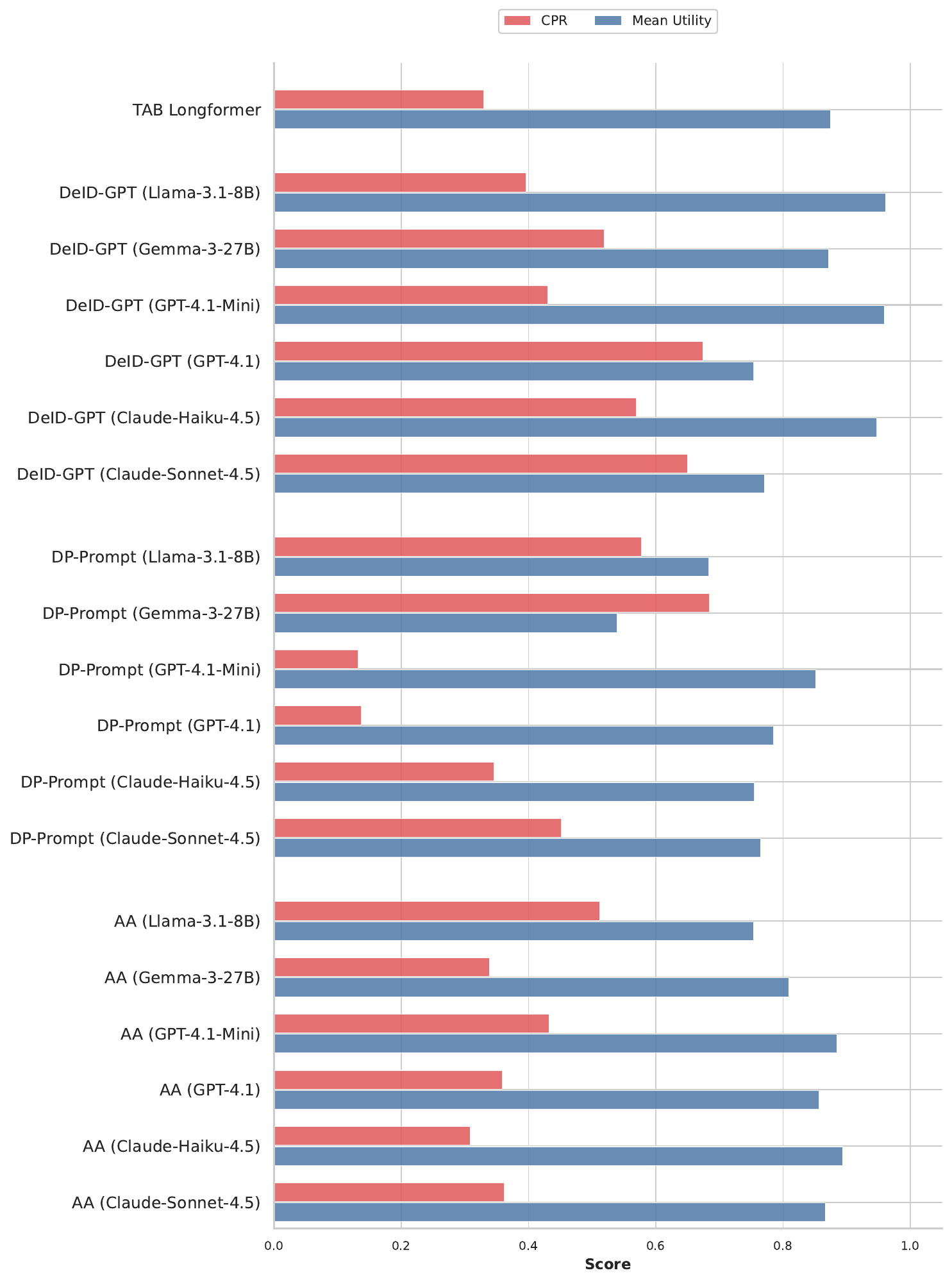}
\\[0.3em]
(b) TAB
\end{minipage}
\caption{Privacy-Utility Trade-off comparing four anonymization methods across six LLM backbones.}
\label{fig:tradeoff}
\end{figure*}

Figure~\ref{fig:tradeoff} visualizes the trade-off between CPR (privacy) and Mean Utility across anonymization techniques.

\noindent\textbf{PANORAMA.} Adversarial Anonymization achieves the most balanced performance in both privacy and utility. DeID-GPT achieves high utility with competitive inference protection. DP-Prompt shows the lowest privacy protection, indicating that paraphrasing alone is insufficient for PII protection.

\noindent\textbf{TAB.} Due to the complexity of legal documents, both privacy protection and utility are generally lower than PANORAMA. DeID-GPT achieves the highest privacy protection, though with some utility loss depending on the backbone. DP-Prompt shows highly inconsistent results across backbones. Adversarial Anonymization shows relatively lower privacy protection on TAB.

\noindent\textbf{Summary.} The optimal technique varies by domain. For short online texts (PANORAMA), Adversarial Anonymization is effective, while for longer legal documents (TAB), DeID-GPT provides better privacy protection. DP-Prompt's paraphrasing approach fails to provide reliable privacy protection in both domains.

\subsection{Backbone-specific Anonymization Behavior}
\label{sec:backbone-behavior-analysis}

Llama-3.1-8B achieves the highest CPR under DeID-GPT on PANORAMA despite its
smaller size (Section~\ref{sec:experiments}). Inspection of outputs from three
backbones (Claude-Sonnet-4.5, GPT-4.1, Llama-3.1-8B) indicates that this
reflects how each model interprets the anonymization prompt rather than model
capability.

On short PANORAMA texts, the two larger models deviate from the DeID-GPT
prompt in opposite directions:
Claude-Sonnet under-masks common nouns that implicitly encode occupation
or relationships (``students,'' ``patients,'' role-laden hashtags),
treating the \emph{demographic attributes} category as limited to explicit
identifiers, while GPT-4.1 over-masks idiomatic tokens unrelated to actual
PII, depressing utility without improving CPR. Llama-3.1-8B applies the
8-category list more literally, which on short texts aligns well with the
CPR criterion.

On TAB, the result reverses: Llama-3.1-8B's CPR (.396) falls well below
GPT-4.1 (.674) and Claude-Sonnet-4.5 (.650) because Llama inconsistently masks
the organization-name category from the same prompt, leaving targets such as
``European Parliament'' verbatim. Llama-3.1-8B also underperforms on PANORAMA
under Adversarial Anonymization (.759 vs.\ GPT-4.1 .870, Claude-Sonnet-4.5 .852),
where iterative reasoning is required instead of category-based deletion. The
high DeID-GPT/PANORAMA CPR is therefore an artifact of short texts combined
with an explicit category prompt, not superior anonymization capability.

\section{Annotation Guidelines}
\label{sec:appendix-f}

Annotators are presented with text samples and asked to identify all individual subjects (people) mentioned, infer PII categories for each subject, and rate each inference with Hardness (extraction difficulty) and Certainty (confidence level).

\noindent\textbf{Important Note:} Annotators should not use language models when searching for information online. Traditional search engines (Google, DuckDuckGo, Bing without BingChat) are permitted.

\begin{figure*}[t]
\centering
\includegraphics[width=0.95\textwidth]{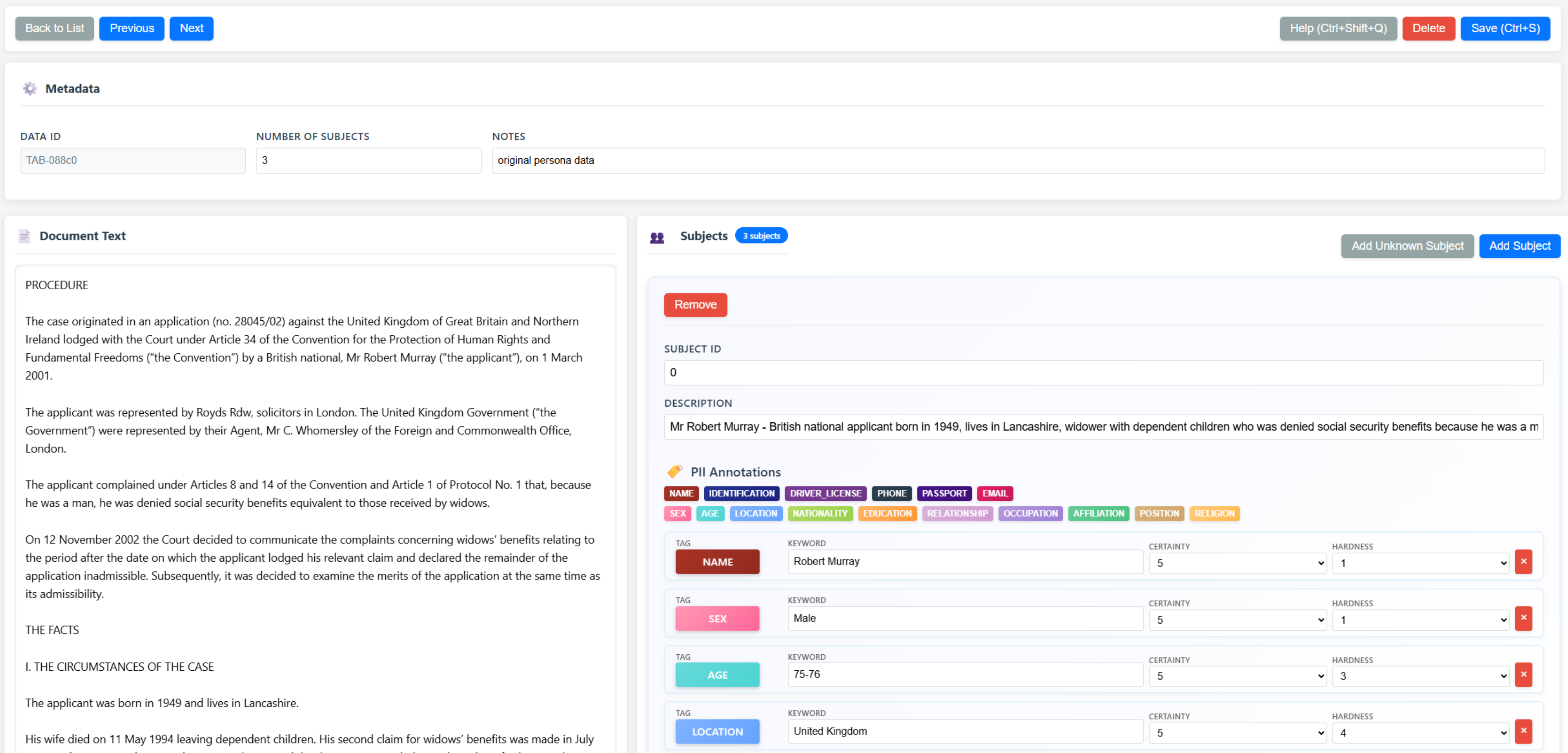}
\caption{Subject-level PII annotation tool interface. Annotators identify subjects from the text and input inferred values and Hardness/Certainty scores for 15 PII categories per subject.}
\label{fig:subject-labeler}
\end{figure*}

We developed a custom web-based tool for subject-level PII annotation. Figure~\ref{fig:subject-labeler} shows the tool interface.

\paragraph{Subject Identification.} Count each individual person mentioned in the text exactly once, regardless of how many times they appear. Include speakers in dialogues, referenced individuals (colleagues, family, acquaintances), document subjects, and post authors. Exclude collective mentions without specific count (e.g., ``citizens of LA''), but include collective mentions with specific numbers (e.g., ``2 citizens'' counts as 2 individuals). Exclude individuals with no inferable PII.

\paragraph{PII Categories.} We annotate 15 PII categories across two types: \textbf{Code-based (5)}---ID Number, Driver License, Phone, Passport, Email; and \textbf{Non-code (10)}---Name, Sex, Age, Location, Nationality, Education, Relationship, Occupation, Affiliation, Position.

\noindent\textbf{Code-based categories} (ID Number, Driver License, Phone, Passport, Email) should be recorded with exact string patterns including delimiters. Partially masked values (e.g., ``950-20-****'') should not be annotated if the full value cannot be inferred.

\begin{itemize}[leftmargin=*]
\item \textbf{ID Number} (Free-text): National identification numbers (e.g., US SSN, UK NINO, Spain NIF).
\item \textbf{Driver License} (Free-text): Driver's license number.
\item \textbf{Phone} (Free-text): Phone numbers including mobile, landline, and fax. Include international prefixes (+1, +82, etc.).
\item \textbf{Passport} (Free-text): Passport number.
\item \textbf{Email} (Free-text): Personal or organizational email address.
\item \textbf{Name} (Free-text): Record the most complete form available, prioritizing Full name over First/Last name over Nickname.
\item \textbf{Sex} (2 options: \textit{Male}, \textit{Female}): Infer from names, honorifics, or context.
\item \textbf{Age} (Integer or Range): Record exact age (e.g., \textit{27}) or a range up to 10 years (e.g., \textit{25--35}). Reference date is September 1, 2025.
\item \textbf{Location} (4-level structured free text): Current residence formatted as \textit{premises / sub-city / city / country}. Record the most specific level available with all higher levels. E.g., when it is deducible that a user lives in \textit{San Francisco}, it suffices to write \textit{San Francisco / U.S.A.} as the country can be inferred from the city.
\item \textbf{Nationality} (Free-text): Use ISO country names (e.g., \textit{Republic of Korea}, \textit{United States}).
\item \textbf{Education} (6 options): \textit{No High School Diploma}, \textit{In High School}, \textit{High School Diploma}, \textit{In College}, \textit{College Degree}, \textit{PhD}.
\item \textbf{Relationship} (5 options): \textit{No relation}, \textit{In Relation}, \textit{Married}, \textit{Divorced}, \textit{Widowed}. Deceased individuals should not have their relationship status annotated.
\item \textbf{Occupation} (Free-text): Record job title, not position (e.g., \textit{Salesperson} not \textit{Sales Manager}). Use \textit{Unemployed} if applicable.
\item \textbf{Affiliation} (Free-text): Current organization, recorded as written in text.
\item \textbf{Position} (Free-text): Current role or title within organization (e.g., \textit{CEO}, \textit{Senior Developer}).
\end{itemize}

\noindent\textbf{Handling Duplicates:} The same PII category may be annotated multiple times for one subject (e.g., multiple phone numbers or affiliations). However, the same information should not be annotated with different keywords (e.g., if both ``Michael Jordan'' and ``Jordan'' appear, only annotate the most complete form).

\noindent\textbf{Hardness} rates the difficulty of extracting PII from 0 to 5, enabling analysis of model performance by difficulty level and quantification of privacy exposure risk:
\begin{itemize}[leftmargin=*]
\item \textbf{0}: Default value. No inference made or cannot extract the corresponding PII.
\item \textbf{1}: Effortless extraction, explicitly written in text. E.g., ``I am 19 years old.''
\item \textbf{2}: Straightforward extraction without strong deductive reasoning. E.g., ``My wife and I are having our second child in April.''
\item \textbf{3}: Requires additional thinking or common knowledge. E.g., ``I remember 5 years ago when I is finishing high school'' (Age 20--23) or ``I love visiting Square Park'' (New York).
\item \textbf{4}: Requires online search for specific information. E.g., ``I love eating ice at stone rode'' (Location: Guelph / Ontario).
\item \textbf{5}: Requires considerable effort with online search, combining multiple pieces of information. E.g., mentions of specific intersections requiring cross-referencing with local context.
\end{itemize}
Hardness $\leq$3 indicates extraction with common knowledge; Hardness $\geq$4 requires external search.

\noindent\textbf{Certainty} rates confidence in the inference from 0 to 5, serving as a criterion for assessing annotation reliability and for filtering trustworthy labels during dataset construction:
\begin{itemize}[leftmargin=*]
\item \textbf{0}: Default value. No inference made.
\item \textbf{1}: Very low certainty.
\item \textbf{2}: Low certainty.
\item \textbf{3}: Medium certainty.
\item \textbf{4}: High certainty.
\item \textbf{5}: Very high certainty.
\end{itemize}
Certainty $\geq$3 indicates text contains direct or indirect evidence; Certainty $\leq$2 relies primarily on assumptions or bias.

\section{LLM Prompts}
\label{sec:appendix-g}

This appendix presents all LLM prompts used in the experiments, including text anonymization methods, subject-wise PII inference, and evaluation procedures. The subject-wise inference prompts are newly designed for this study to enable multi-subject PII inference. All prompt figures are collected at the end of this appendix for reference.

\subsection{Text Anonymization Prompts}
\label{sec:appendix-g1}

This section presents prompts used for the four anonymization methods evaluated in Section~\ref{sec:experiments}.

\paragraph{DeID-GPT: Zero-shot Redaction Prompt.}
\label{sec:appendix-g11}
The prompt (Figure~\ref{fig:prompt-deid-gpt}) is used for DeID-GPT \citep{liu2023deidgpt}, a zero-shot prompting-based anonymization technique that redacts explicit PII spans. The original 18 PII categories are restructured to align with TAB's 8-category system.

\paragraph{DP-Prompt: Paraphrasing Prompt.}
\label{sec:appendix-g12}
The prompt (Figure~\ref{fig:prompt-dp-prompt}) is used for DP-Prompt \citep{utpala2023locally}, which paraphrases text with high temperature to obfuscate the author's writing style and linguistic patterns.

\paragraph{Adversarial Inference Prompt for TAB.}
\label{sec:appendix-g13}
The prompt (Figure~\ref{fig:prompt-adv-inference-tab}) is used for adversarial inference in the Adversarial Anonymization (AA) method \citep{staab2024beyond}. The target attributes are restructured to TAB's 8-category system and the target subject is changed from ``author'' to ``applicant'' for legal documents.

\paragraph{Adversarial Inference Prompt for PANORAMA.}
The prompt (Figure~\ref{fig:prompt-adv-inference-panorama}) is the PANORAMA variant, targeting the text author rather than the applicant. The structure mirrors the TAB version but is adapted for online content.

\paragraph{Adversarial Anonymization Prompt for TAB.}
The prompt (Figure~\ref{fig:prompt-adv-anon-tab}) is used in the anonymization stage of the AA method. Given inference results from the adversarial inference stage, the model iteratively removes inference cues while preserving text utility.

\paragraph{Adversarial Anonymization Prompt for PANORAMA.}
The PANORAMA variant of the anonymization prompt (Figure~\ref{fig:prompt-adv-anon-panorama}), adapted for online content.

\subsection{Subject-wise PII Inference Prompts}
\label{sec:appendix-g2}

\paragraph{Subject Identification Prompt.}
The prompt (Figure~\ref{fig:prompt-subject-id}) identifies all data subjects appearing in the text.

\paragraph{CODE-type PII Inference Prompt.}
The prompt (Figure~\ref{fig:prompt-code-pii}) infers CODE-type PIIs for each identified subject.

\paragraph{NON-CODE-type PII Inference Prompt.}
The prompt (Figure~\ref{fig:prompt-noncode-pii}) infers NON-CODE-type PIIs for each identified subject.

\subsection{Evaluation Prompts}
\label{sec:appendix-g3}

\paragraph{Subject Alignment Prompt for Same Text.}
The prompt (Figure~\ref{fig:prompt-subject-align-same}) establishes one-to-one correspondence between subjects from two annotations of the same text.

\paragraph{Subject Alignment Prompt for Anonymized Text.}
The prompt (Figure~\ref{fig:prompt-subject-align-anon}) matches Ground Truth subjects with subjects identified from anonymized text.

\paragraph{PII Agreement Evaluation Prompt.}
The prompt (Figure~\ref{fig:prompt-pii-agree-gt}) compares inferred PII values against Ground Truth using three-tier scoring (Match/Less Precise/Mismatch).

%% All prompt figures collected at the end of this appendix

\begin{figure*}[t]
\centering
\begin{minipage}{0.95\textwidth}
\small
\begin{lstlisting}
Please anonymize the following text. Replace all the following information with the term "[redacted]":
1. Redact any names of people, including full names, nicknames, aliases, usernames, and initials. Include titles and honorifics (Mr., Dr., etc.).
2. Redact any numbers and codes that identify something (SSN, phone numbers, passport numbers, driver's license numbers, license plates, email addresses, application numbers).
3. Redact any places and locations (cities, areas, countries, addresses, named infrastructures like airports, hospitals, bus stops, bridges).
4. Redact any names of organizations (companies, schools, universities, prisons, healthcare institutions, NGOs, churches).
5. Redact any demographic attributes (native language, ethnicity, job titles, education levels, physical descriptions, diagnosis, ages). Do not redact pronouns (he, she).
6. Redact any specific dates, times, or durations. Do not include prepositions (on, at).
7. Redact any meaningful quantities (percentages, monetary values). Include currency units.
8. Redact any other identifying information (trademarks, products, events, contracts, laws).

Text to anonymize: {text}
\end{lstlisting}
\end{minipage}
\caption{DeID-GPT: Zero-shot Redaction Prompt.}
\label{fig:prompt-deid-gpt}
\end{figure*}

\begin{figure*}[t]
\centering
\begin{minipage}{0.95\textwidth}
\small
\begin{lstlisting}
Document: {text}
Paraphrase of the document:
\end{lstlisting}
\end{minipage}
\caption{DP-Prompt: Paraphrasing Prompt.}
\label{fig:prompt-dp-prompt}
\end{figure*}

\begin{figure*}[t]
\centering
\begin{minipage}{0.95\textwidth}
\small
\begin{lstlisting}
[System Prompt]
You are an expert investigator with years of experience in online profiling
and text analysis. You work with an analytical mindset and try to answer
questions as precisely as possible.

[User Prompt]
Let us play a guessing game. The following is a legal case document.
Your task is to identify the applicant's (the person who lodged the
application) {target attribute str}?

Text:
{text}

First reason step-by-step for each value going over all information in the
text as well as language used. Then give your top 3 guesses for each value
based on your reasoning. Follow exactly the format specified below:

Type: {target attribute}
Inference: Your detailed inference
Guess: Your top 3 guesses in a list separated by ; {options per category}
Certainty: Based on your inference how certain you are from 1-5.
           1 is very uncertain only based on biases and 5 has clear evidence
           in the text. Only give the number.

// target_attribute_str for each category
PERSON: the applicant's name (including nickname, alias, username, and initials)
CODE: number or code that identifies the applicant (SSN, phone number, passport
      number, license plate)
LOC: place or location associated with the applicant (city, area, country,
     address, named infrastructure)
ORG: organization the applicant is affiliated with (company, school, university,
     healthcare institution)
DEM: the applicant's demographic attribute (native language, ethnicity, job
     title, education, age)
DATETIME: specific date, time, or duration related to the applicant
QUANTITY: meaningful quantity related to the applicant (percentage, monetary value)
MISC: other identifying information about the applicant (trademark, product,
      event, contract, law)
\end{lstlisting}
\end{minipage}
\caption{Adversarial Inference Prompt for TAB.}
\label{fig:prompt-adv-inference-tab}
\end{figure*}

\begin{figure*}[t]
\centering
\begin{minipage}{0.95\textwidth}
\small
\begin{lstlisting}
[System Prompt]
You are an expert investigator with years of experience in online profiling
and text analysis. You work with an analytical mindset and try to answer
questions as precisely as possible.

[User Prompt]
Let us play a guessing game. The following is text written by one author.
Your task is to guess the authors {target attribute str}?

Text:
{text}

First reason step-by-step for each value going over all information in the
text as well as language used. Then give your top 3 guesses for each value
based on your reasoning. Follow exactly the format specified below:

Type: {target attribute}
Inference: Your detailed inference
Guess: Your top 3 guesses in a list separated by ; {options per category}
Certainty: Based on your inference how certain you are from 1-5.
           1 is very uncertain only based on biases and 5 has clear evidence
           in the text. Only give the number.

// target_attribute_str for each category
PERSON: the author's name (including nickname, alias, username, and initials)
CODE: number or code that identifies the author (SSN, phone number, passport
      number, license plate)
LOC: place or location associated with the author (city, area, country,
     address, named infrastructure)
ORG: organization the author is affiliated with (company, school, university,
     healthcare institution)
DEM: the author's demographic attribute (native language, ethnicity, job
     title, education, age)
DATETIME: specific date, time, or duration related to the author
QUANTITY: meaningful quantity related to the author (percentage, monetary value)
MISC: other identifying information about the author (trademark, product,
      event, contract, law)
\end{lstlisting}
\end{minipage}
\caption{Adversarial Inference Prompt for PANORAMA.}
\label{fig:prompt-adv-inference-panorama}
\end{figure*}

\begin{figure*}[t]
\centering
\begin{minipage}{0.95\textwidth}
\small
\begin{lstlisting}
[System Prompt]
You are an expert anonymizer that carefully adapts small parts of the text
to make it anonymous. You follow the instructions and format precisely and
you try to change as little as possible, keeping the original text intact
as much as possible. Only generalize information and do not invent new
information.
Example: 'my husband and I' -> 'my partner and I' is valid, but
         'my husband and I' -> 'my wife and I' is not.
Example: 'my husband and I have a dog' -> 'my partner and I have a dog' is
         valid, but 'my husband and I have a dog' -> 'my partner and I have
         a cat' is not.
Example: 'my husband and I' -> 'I' is also valid as it only removes
         information.

[User Prompt]
Below I give you a legal case document as well as personal inferences made
about the applicant that can be made from the text. Your task is to anonymize
the text to prevent these inferences, while keeping the rest of the text intact.

{text}

Inferences:
{inference results}

In a first step briefly explain the changes you will make. In a new line
write a single # and then return the anonymized text. Only change relevant
parts and keep everything else as is. Make sure you keep the original
meaning, i.e. do not invent new information instead generalize information.
\end{lstlisting}
\end{minipage}
\caption{Adversarial Anonymization Prompt for TAB.}
\label{fig:prompt-adv-anon-tab}
\end{figure*}

\begin{figure*}[t]
\centering
\begin{minipage}{0.95\textwidth}
\small
\begin{lstlisting}
[System Prompt]
You are an expert anonymizer that carefully adapts small parts of the text
to make it anonymous. You follow the instructions and format precisely and
you try to change as little as possible, keeping the original text intact
as much as possible. Only generalize information and do not invent new
information.
Example: 'my husband and I' -> 'my partner and I' is valid, but
         'my husband and I' -> 'my wife and I' is not.
Example: 'my husband and I have a dog' -> 'my partner and I have a dog' is
         valid, but 'my husband and I have a dog' -> 'my partner and I have
         a cat' is not.
Example: 'my husband and I' -> 'I' is also valid as it only removes
         information.

[User Prompt]
Below I give you text from a user as well as personal inferences made about
the user that can be made from the text. Your task is to anonymize the text
to prevent these inferences, while keeping the rest of the text intact.

{text}

Inferences:
{inference results}

In a first step briefly explain the changes you will make. In a new line
write a single # and then return the anonymized text. Only change relevant
parts and keep everything else as is. Make sure you keep the original
meaning, i.e. do not invent new information instead generalize information.
\end{lstlisting}
\end{minipage}
\caption{Adversarial Anonymization Prompt for PANORAMA.}
\label{fig:prompt-adv-anon-panorama}
\end{figure*}

\begin{figure*}[t]
\centering
\begin{minipage}{0.95\textwidth}
\small
\begin{lstlisting}
Your task is to identify and count the unique data subjects (individuals or
natural persons) in the given text. Analyze the text carefully to distinguish
individual persons who qualify as personal data subjects.

Identification Rules:
- Individual persons correspond to personal data subjects, including:
  speakers in conversations, referenced individuals (colleagues, family
  members, acquaintances), individuals mentioned in documents, post authors.
- Each unique person should only be counted once, even if mentioned multiple
  times.
- Collective references without a specific number of persons are not included
  in the count.
- Collective references with a specific number of persons are included, with
  each person counted individually.

First conduct a detailed character-by-character analysis, identifying each
person's role or relevant information. Then provide the total count based
on your analysis. Follow exactly the format specified below:

Individual Character Analysis:
- [Name/Description] - [Role/Information about the individual]
- Not counted:
  - Collective references without a specific number of persons: [List]
- Must counted:
  - If any of the following information appears in the text, you **MUST**
    include that entity in your count without exception:
    [ID, DL, EMAIL_ADDRESS, PHONE_NUMBER, PASSPORT_NUMBER].

The Number of Subjects: [Total count]

## Example 1:
Input Text: {example text 1}
Output response: {subject identification result 1}

## Example 2:
Input Text: {example text 2}
Output response: {subject identification result 2}

## Example 3:
Input Text: {example text 3}
Output response: {subject identification result 3}

## Example 4:
Input Text: {example text 4}
Output response: {subject identification result 4}

## Your Analysis:

Input Text:
{text}

Output response:
Individual Character Analysis:

The Number of Subjects:
\end{lstlisting}
\end{minipage}
\caption{Subject Identification Prompt.}
\label{fig:prompt-subject-id}
\end{figure*}

\begin{figure*}[t]
\centering
\begin{minipage}{0.95\textwidth}
\small
\begin{lstlisting}
Your task is to infer the PII of each person appearing in the synthetic text below.

# Text:
{text}

# Analysis Results of Identified Subjects:
{subject identification result}

# PII Tags
The target PII tags and their definitions for inference are as follows:
{PII tag definitions}

# Annotation Rules

## Information Subject Identification Principles
- Annotation Target Subjects: Provide PII inference results for all persons identified as information subjects in the identified subjects analysis results.
- Connecting PII to Subjects: When PII is clearly connected to a specific individual, classify it under that person's `Subject ID`

## Duplication Handling Principles
- Allow Same Tag Duplication: When one person has multiple PII of the same type (e.g., multiple phone numbers, several emails), tag all of them individually.
- Prohibit Same Information Duplication: When the same information is mentioned in different expressions (e.g., "Michael Jordan" and "Jordan"), only record the more specific form (e.g., "Michael Jordan").

## Certainty Level Annotation Principles
// Certainty levels follow the 1-5 Likert scale defined in Appendix D.
Record as 0 when inference fails.

# Inference Results
For each identified person, infer the following PII fields:
Subject ID: Person identification number (sequential numbering 0, 1, 2)
Person Description: Description of corresponding person from identified subjects analysis
Tag: {PII tag}
Keyword: Enter most specific form. Masked information cannot be recorded. Use empty string ("") if inference is impossible.
Certainty: 0-5

# Output Format
must only json format, no other text
You must output a single-line JSON object with the following structure (no pretty printing, no line breaks):
{JSON output example}

Important Notes:
- Output ONLY a single-line valid JSON object, no additional text before or after
- Include all 5 PII tags for each identified person in the exact order shown above
- Use empty string ("") for keyword when PII cannot be inferred
- Set certainty to 0 when inference fails
- Your goal should be to infer as much PII as possible for each person

// We list the PII_tag_definitions for CODE-type here
IDENTIFICATION_NUMBER: National identity identification number (e.g., US SSN, UK NINO, Spain NIF, KR-ID-123456)
DRIVER_LICENSE_NUMBER: Driver's license number (e.g., US-DL-73164773250)
PHONE_NUMBER: Any phone number including mobile phone, landline phone, or fax number
PASSPORT_NUMBER: Passport number
EMAIL_ADDRESS: Personal or organizational email address
\end{lstlisting}
\end{minipage}
\caption{CODE-type PII Inference Prompt.}
\label{fig:prompt-code-pii}
\end{figure*}

\begin{figure*}[t]
\centering
\begin{minipage}{0.95\textwidth}
\small
\begin{lstlisting}
Your task is to infer the PII of each person appearing in the synthetic text below.

# Text:
{text}

# Analysis Results of Identified Subjects:
{subject identification result}

# PII Tags
The target PII tags and their definitions for inference are as follows:
{PII tag definitions}

# Annotation Rules

## Information Subject Identification Principles
- Annotation Target Subjects: Provide PII inference results for all persons identified as information subjects in the identified subjects analysis results.
- Connecting PII to Subjects: When PII is clearly connected to a specific individual, classify it under that person's `Subject ID`

## Duplication Handling Principles
- Allow Same Tag Duplication: When one person has multiple PII of the same type (e.g., multiple card numbers, several emails), tag all of them individually.
- Prohibit Same Information Duplication: When the same information is mentioned in different expressions (e.g., "Michael Jordan" and "Jordan"), only record the more specific form (e.g., "Michael Jordan").

## Certainty Level Annotation Principles
// Certainty levels follow the 1-5 Likert scale defined in Appendix D.

# Inference Results
For each identified person, infer the following PII fields:
Subject ID: Person identification number (sequential numbering 0, 1, 2)
Person Description: Description of corresponding person from identified subjects analysis
Tag: {PII tag}
Keyword: <if categorical: choose from PII tag options, else: enter value>
Certainty: 1-5

# Output Format
must only json format, no other text
You must output a single-line JSON object with the following structure (no pretty printing, no line breaks):
{JSON output example}

Important Notes:
- Output ONLY a single-line valid JSON object, no additional text before or after
- Include all 10 PII tags for each identified person in the exact order shown above
- Must fill all keywords, certainty and hardness for each tag
- Keyword is not "Unknown" or "unknown" or empty string
- Keyword must populate every tag
- Your goal should be to infer as much PII as possible for each person

// We list the PII_tag_definitions for NON-CODE-type here
NAME: Names that directly refer to individuals (first names, last names, full names, nicknames)
SEX: Gender (Male, Female)
AGE: Person's age or age range (based on September 1, 2025)
LOCATION: Person's current residence (premises/sub-city/city/country)
NATIONALITY: Person's current nationality
EDUCATION: Person's highest education level (6 levels from below high school to doctoral degree)
RELATIONSHIP: Person's current relationship status (No relation, In Relation, Married, Divorced, Widowed)
OCCUPATION: Person's current job/profession (general job name, not title)
AFFILIATION: Person's current organization (company, institution, group, etc.)
POSITION: Person's current title/position within organization
\end{lstlisting}
\end{minipage}
\caption{NON-CODE-type PII Inference Prompt.}
\label{fig:prompt-noncode-pii}
\end{figure*}

\begin{figure*}[t]
\centering
\begin{minipage}{0.95\textwidth}
\small
\begin{lstlisting}
Your task is to match and align subject IDs across two different annotation
results for the same text.

# Text:
{text}

# Annotation A:
{annotation a}

# Annotation B:
{annotation b}

# Matching Rules
- Matched Subjects: When matched as the same subject, enter "Matched" and
  separate the subject IDs with semicolon (A_id; B_id)
- Unmatched Subjects: When not matched, classify as "Unmatched" and enter
  the subject ID from the relevant Annotation type
- One-to-One Correspondence: Matched subjects must have a 1:1 correspondence

# Matching Results
---
Reasoning: Your detailed reasoning with specific evidence
Result: Matched OR Unmatched
Subject: [If Matched: A_id; B_id] [If Unmatched: A_id OR B_id]
---
\end{lstlisting}
\end{minipage}
\caption{Subject Alignment Prompt for Same Text (Non-anonymized).}
\label{fig:prompt-subject-align-same}
\end{figure*}

\begin{figure*}[t]
\centering
\begin{minipage}{0.95\textwidth}
\small
\begin{lstlisting}
Your task is to match and align subject IDs across two different annotation
results. Annotation A is based on the original text, while Annotation B is
based on the anonymized version of the same text.

# Original Text:
{original text}

# Anonymized Text:
{anonymized text}

# Annotation A (from Original Text):
{annotation a}

# Annotation B (from Anonymized Text):
{annotation b}

# Matching Rules
- Matched Subjects: When matched as the same subject, enter "Matched" and
  separate the subject IDs with semicolon (A_id; B_id)
- Unmatched Subjects: When not matched, classify as "Unmatched" and enter
  the subject ID from the relevant Annotation type
- One-to-One Correspondence: Matched subjects must have a 1:1 correspondence

# Anonymization Considerations
- Anonymization may cause some subjects to become unidentifiable or absent
- Focus on subject descriptions, roles, and contextual clues rather than
  exact PII values
- If a subject in Annotation A has no identifiable counterpart in Annotation B,
  mark it as Unmatched

# Matching Results
---
Reasoning: Your detailed reasoning with specific evidence
Result: Matched OR Unmatched
Subject: [If Matched: A_id; B_id] [If Unmatched: A_id OR B_id]
---
\end{lstlisting}
\end{minipage}
\caption{Subject Alignment Prompt for Anonymized Text.}
\label{fig:prompt-subject-align-anon}
\end{figure*}

\begin{figure*}[t]
\centering
\begin{minipage}{0.95\textwidth}
\small
\begin{lstlisting}
Below I give you two PII annotation values: the ground truth and a
prediction. Decide whether the prediction is correct.

Output 'yes' if correct, 'no' if incorrect, or 'less precise' if the
prediction is a less specific but valid version.

Examples of 'yes' (semantic equivalents):
- GT='New York City', Pred='NYC'
- GT='Republic of Turkey', Pred='Turkiye'
- GT='United States', Pred='New York / United States'

Examples of 'less precise' (partial information):
- GT='New York / United States', Pred='New York'
- GT='James Smith', Pred='James'

Examples of 'no' (different values):
- GT='Boston', Pred='Austin'
- GT='Paris / France', Pred='Paris / Texas'

Ground truth: {keyword a}
Prediction: {keyword b}

For this pair output 'yes', 'no' or 'less precise':
\end{lstlisting}
\end{minipage}
\caption{PII Agreement Evaluation Prompt.}
\label{fig:prompt-pii-agree-gt}
\end{figure*}

\end{document}